\documentclass[final]{siamonline250211}
\usepackage{amsmath, amssymb}
\usepackage{amsmath,amssymb}
\newsiamremark{remark}{Remark}

\usepackage{xr-hyper}

\usepackage{booktabs}
\usepackage{caption}
\usepackage{float}
\usepackage{placeins}
\newtheorem{example}[theorem]{Example}

\title{A Constructive Orbit-Based Theory of Learnability}

\headers{A Constructive Orbit-Based Theory of Learnability}{Y. Puig de Dios}

\author{
Yunied Puig\thanks{
Email: \email{puigdedios@gmail.com}.
}
}

\begin{document}
\maketitle


\begin{abstract}
We develop a finite-dimensional orbit framework for constructing data-adapted dictionaries, sparse realizations, and predictive extensions. The method uses wrapped weighted shifts to generate orbit dictionaries and local interpolation maps from independent orbit representatives, turning the learning problem into three explicit tasks: scaffold discovery, sparse realization on the discovered dictionary, and prediction through learned sparse coordinates. We introduce CCDS as the main constructive ingredient: a deviation-controlled sparse realization procedure that recovers MP/OMP-type behavior at the minimal-deviation boundary while allowing feasible interior paths with prescribed residual geometry that is inaccessible to MP/OMP. We prove generic orbit independence, exact block interpolation with norm
control, observable bounds on sparse approximation distortion, sparse
realization guarantees on selected supports, and fixed-path stability
under admissible perturbations of the block interpolants. These results
connect sparse approximation, interpolation, and finite-dimensional
operator dynamics through the construction and analysis of data-induced
orbit dictionaries. Experiments on synthetic geometry, controlled realization tests, and standard classification benchmarks show that selected orbit scaffolds have substantially better matched-rank projection behavior than random orbit or unstructured alternatives, that CCDS attains near-zero realization error at the minimal-deviation boundary, and that predictors trained from the sparse CCDS certificates give competitive performance.
\end{abstract}



\section{Introduction}

Modern learning systems achieve remarkable predictive performance, but their internal mechanisms are often difficult to characterize mathematically. In this work we study a different organizing principle for learning: rather than beginning with unrestricted parametrization, we begin with a structured orbit geometry induced by the data and ask what targets can be constructively realized on it. This leads to an orbit-based framework organized around three linked bottlenecks: a \emph{scaffold bottleneck}, measuring how well a data-induced orbit scaffold captures the relevant target geometry; a \emph{realizability bottleneck}, measuring how accurately the projected target can be synthesized on that scaffold; and a \emph{predictive coefficient bottleneck}, measuring how well the resulting constructive coefficient rule can be extended to unseen inputs.

The main contribution is a constructive controlled-deviation
synthesis mechanism (CCDS) on data-induced orbit dictionaries.
Wrapped weighted shifts and block interpolation provide the dictionary;
CCDS then characterizes the feasible stepwise deviations and constructs
the corresponding sparse realizations. Minimal-deviation choices recover
MP and OMP, while nonzero interior updates departing from an OMP
projection state cannot be produced by OMP under any atom ordering.
For prescribed interior paths with positive feasibility margins, the
realization is locally Lipschitz stable under admissible perturbations
of the block interpolants. An observable local-chart score further
controls the distortion of bounded-coefficient sparse approximation,
providing a geometric justification for dictionary pre-selection.

The predictive model, which we call a \emph{Weighted Backward Shift Neural Network} (WBSNN), is built directly on this constructive structure. Phase~1 identifies the orbit scaffold, Phase~2 uses CCDS to associate each training target with a sparse coefficient certificate over the fixed orbit dictionary, and Phase~3 learns an extension of this coefficient map to unseen inputs. Thus prediction is not introduced as an independent parametrization problem: it extends the constructive rule identified by the first two phases. Nonlinearity enters through the data-dependent selection and combination of orbit-generated atoms and CCDS paths, while prediction further extends this structure through a learned coefficient map to unseen inputs.

This viewpoint provides both structural interpretability and practical flexibility. Every realization and prediction is expressed through coefficients over a fixed orbit-derived representation, while the projection and parametrization errors expose where the construction succeeds or fails. The theory also shows that exact block interpolation is not required in practice: regularized block interpolants preserve the scaffold geometry locally and perturb the CCDS realization continuously, justifying a better-conditioned relaxed implementation. Empirically, across synthetic and real datasets, the framework produces highly sparse realizations and competitive predictive performance; in several experiments the relevant orbit scaffold is recovered using only a small fraction, as little as \(1\%\), of the training data.

The construction is rooted in linear dynamics. Weighted backward shifts provide a simple operator class whose iterates generate rich orbit structure, and we introduce a wrapped finite-dimensional version that preserves this mechanism while avoiding coordinate loss. Its iterates generate the dictionaries on which scaffold construction, constructive realization, and prediction are built. The resulting perspective shifts attention from parametrization alone to the interaction between dataset geometry, orbit structure, and constructive realizability.

The paper is organized as follows. Our main results are in Section \ref{sec:realizability}, and experimental results are in Section \ref{sec:experiments}. Appendix~\ref{app:proofs and tables} contains the complete proofs, while
Appendix~\ref{sec:experiment-config} records the experimental configurations.

\section{Relation to prior work}
Our deviation-control viewpoint through CCDS is motivated by Bernstein lethargy
theory, where approximation errors along nested families need not obey 
a universal decay rate \cite{Bernstein1938,Shapiro1964}. Rather than
postulating such a rate, we treat the residual trajectory itself as a
constructive object. This differs from discrepancy principles, where a
prescribed residual typically serves as a stopping or regularization
criterion \cite{Morozov1966,Bakushinsky2004}, and from classical
greedy pursuits such as MP and OMP, whose residuals are determined by
greedy contraction or orthogonal projection
\cite{Mallat1993,Pati1993,Tropp2007,DeVore1996,Temlyakov2011}.
In our framework, feasible residual profiles are prescribed and realized
explicitly; standard greedy updates arise as boundary regimes of this
broader deviation-control mechanism.

Classical dictionary-learning methods, including
K-SVD~\cite{aharon2006ksvd} and online dictionary
learning~\cite{mairal2010online}, adapt dictionary atoms to data
through sparsity-constrained or sparsity-penalized reconstruction.
Our construction instead generates atoms through the structured
family \(J_kW^m x_{k,j}\), coupling dictionary geometry to anchor
interpolation. On the resulting dictionary, CCDS constructs sparse
realizations with prescribed feasible deviations and fixed-path stability under admissible perturbations, recovering MP/OMP
at the minimal-deviation boundary.

Operator-theoretic ideas in machine learning also appear in
Koopman-based representations and forecasting
\cite{Mezic2005,Lusch2018,Azencot2020,Brunton2022} and in
neural-operator approaches such as DeepONet and neural integral
equations \cite{Lu2021,Zappala2024}. WBSNN is different in
that it does not learn an operator representation directly: iterates of
a single orbit-generating operator first induce a data-dependent
scaffold, targets are then constructively realized on that scaffold,
and prediction is obtained by extending the resulting coefficient law.
Its mathematical origin is therefore closer to linear dynamics and
orbit theory, with weighted backward shifts providing the guiding
prototype \cite{Bayart2009}.

\section{Our Orbit-Based Constructive Theory}
\label{sec:realizability}

This section develops the constructive backbone of the paper. 
The central object is a data-induced orbit scaffold $U\subseteq\mathbb{R}^C$ constructed from a small anchor subset of a given dataset. The target is first projected onto this scaffold and then constructively realized within it. This separates realization into two distinct problems: how well the scaffold captures the target on the dataset, and how accurately the projected target can be synthesized once the scaffold is fixed. These will become, respectively, the projection and realization bottlenecks developed below.

\paragraph{Standing data notation}
Throughout, let \(\Xi_{\rm train}=\{(x_i,y_i)\}_{i=1}^n\subset\mathbb{R}^d\times\mathbb R^C\) be a finite input--target dataset generated by an empirical target map \(f_\Xi:\mathbb R^d\to\mathbb R^C\), so that $f_\Xi(x_i):=y_i$.
 We define the empirical mean seminorm for any map \(h\) evaluated on the dataset by
$
\|h\|_{\Xi}:=\frac{1}{n}\sum_{i=1}^n\|h(x_i)\|_2.
$

Let \(\Xi_{\rm sub}\subseteq\Xi_{\rm train}\) be
the anchor subset used for scaffold
construction.
We write
\begin{equation}
\label{eq:blocks def}
\Xi_{\mathrm{sub}}=D_1\sqcup\cdots\sqcup D_K,
\qquad
D_k=\{(x_{k,j},y_{k,j})\}_{j=1}^{s_k}.
\end{equation}

\subsection{The framework's backbone}
Our starting point is the observed input--target relation. Even within
an anchor block, there need not exist a single linear operator sending
every original input to its prescribed target. We therefore seek a
structured representation of these constraints by first transforming
the inputs along operator orbits and then applying a blockwise linear
readout.

Weighted backward shifts are classical operators in linear dynamics and can exhibit remarkably rich orbit behavior in infinite-dimensional spaces. To adapt their orbit mechanism to finite-dimensional learning, we use a wrapped weighted backward shift, which replaces coordinate loss by cyclic propagation modulo $d$. This provides the structured orbit backbone used throughout the framework.

\begin{definition}[The wrapped weighted backward shift]
\label{def:wrapped shift}
A wrapped weighted backward shift $W:\mathbb R^d\to \mathbb R^d$ is defined as  
$We_j=w_je_{j-1\pmod{d}}$,
 where $e_0,\ldots,e_{d-1}$ is the canonical basis and $w_j\neq0$.
\end{definition}

Even within a block, the original assignments
\(x_{k,j}\mapsto y_{k,j}\) need not admit a common linear interpolant.
We therefore use anchor-dependent iterates of \(W\) to seek orbit
representatives on which a shared linear readout can realize the
prescribed targets. Specifically, in the next subsection, for an
anchor block \(D_k\), we seek depths \(L_{k,j}\) and a linear
operator \(J_k\) such that
\[
x_{k,j}
\longmapsto W^{L_{k,j}}x_{k,j}
\longmapsto J_kW^{L_{k,j}}x_{k,j}=y_{k,j}.
\]
The two operators have then distinct roles: \(W\) generates the orbit
geometry, while \(J_k\) converts the selected orbit representatives
into target outputs. The depth may depend on the anchor, although
the readout is shared within each block. These identities realize
the anchor constraints in \(\Xi_{\mathrm{sub}}\).

\begin{lemma}
\label{lem:iterates of wrapped wbs}
For every $m\ge1$ and $z\in\mathbb R^d$,
\[
\bigl(W^m z\bigr)_{i-1}
=
\left(\prod_{r=0}^{m-1} w_{(i+r)\bmod d}\right)
z_{(i+m-1)\bmod d},
\qquad i=1,\ldots,d.
\]
\end{lemma}
The proof is deferred to Subsection~\ref{subsec:wrapped weighted back shifts}.
For intuition, the reader may view \(W^{L}x\) as a shift-and-scale transformation of \(x\), beginning at the \(L\)-th entry and preserving the cyclic order of the components modulo \(d\).

Set $\rho:=\prod_{j=0}^{d-1}w_j$. Since every coordinate crosses every weight exactly once in $d$ iterates, $W^d=\rho I_{\mathbb R^d}$. Hence:
 \begin{lemma}
\label{lem:form orbits of W}
If
$L=md+r$,
$\
0\leq r<d$,
then
$W^L
=
\rho^mW^r.$
\end{lemma}

\subsection{Blockwise orbit interpolation}

Fix a partition as in (\ref{eq:blocks def}) with \(s_k\le d\) for every \(k\).
We call the partition orbit-admissible for \(W\) if, for each
\(1\le k\le K\), there exist depths
\[
L_{k,1},\dots,L_{k,s_k}\in\{0,\dots,d-1\}
\]
such that the orbit representatives
\[
z_{k,j}:=W^{L_{k,j}}x_{k,j},
\qquad
j=1,\dots,s_k,
\]
are linearly independent. In what follows, assume that the partition is
orbit-admissible and choose such depths for each block.

Define
\begin{equation}
\label{eq:matrices Z and Y}
Z_k
:=
\begin{bmatrix}
z_{k,1}&\cdots&z_{k,s_k}
\end{bmatrix}
\in\mathbb{R}^{d\times s_k},
\qquad
Y_k
:=
\begin{bmatrix}
y_{k,1}&\cdots&y_{k,s_k}
\end{bmatrix}
\in\mathbb{R}^{C\times s_k}.
\end{equation}

Since $Z_k$ has full column rank, the canonical interpolant
$
J_k:=Y_kZ_k^\dagger\in\mathcal{L}(\mathbb{R}^d,\mathbb{R}^C)
$
satisfies
\begin{equation}
\label{eq:exact int eq}
J_kW^{L_{k,j}}x_{k,j}
=
J_kz_{k,j}
=
y_{k,j},
\qquad
j=1,\dots,s_k.
\end{equation}
Moreover,
$
\|J_k\|
\leq
\|Y_k\|_F/\sigma_{\min}(Z_k),
$
so the conditioning of the orbit matrix $Z_k$ controls the amplification associated with blockwise interpolation.

Additionally, we show that exact interpolation also survives in the infinite-dimensional setting through a constructive application of the Hahn--Banach theorem; readers interested in this extension are referred to Subsection~\ref{app:exact-inetrp-inf-dim}.

A natural question is how restrictive the orbit-independence hypothesis underlying exact interpolation is. The next result shows that, whenever this condition is feasible for a given block, it is topologically generic and robust, and holds Lebesgue-almost everywhere: the admissible weights form an open dense subset of weight space, and the exceptional set has Lebesgue measure zero. Complete proof in Subsection~\ref{app:exact-inetrp-inf-dim}.

\begin{theorem}[Genericity of the orbit-independence condition]
\label{thm:generic_orbit_independence}
Let
$
D={(x_i,y_i)}_{i=1}^s\subset\mathbb R^d\times\mathbb R^C,
\
s\leq d,
$
and let
$
\Omega:=(\mathbb R\setminus{0})^d.
$
For each
$
w=(w_0,\dots,w_{d-1})\in\Omega,
$
let $W_w$ denote the wrapped weighted backward shift determined by $w$. Define $
z_i:=W_w^{L_i}x_i
$ and

\[
\Omega_D
:=
\Bigl\{
w\in\Omega:
\exists\,L_1,\dots,L_s\in\{0,\dots,d-1\}
\text{ such that }
z_i\notin\operatorname{span}\{z_1,\dots,z_{i-1}\}
\text{ for }1\leq i\leq s
\Bigr\}.
\]

If $\Omega_D\neq\varnothing$, then $\Omega_D$ is open and dense in $\Omega$. Moreover,
$
\Omega\setminus\Omega_D
$
has Lebesgue measure zero.
\end{theorem}

\begin{proof}[Proof sketch]
For each depth tuple \(L=(L_1,\ldots,L_s)\), the entries and
\(s\times s\) minors of
\[
M_L(w)=[W_w^{L_1}x_1\ \cdots\ W_w^{L_s}x_s]
\]
are polynomial in \(w\). Thus \(\Omega_D\) is a finite union of
polynomial nonvanishing sets and is open. If \(\Omega_D\neq\varnothing\),
some minor \(p\) is nonzero at an admissible weight and hence is not
the zero polynomial. Since
\[
\Omega\setminus\Omega_D\subseteq\{w\in\Omega:p(w)=0\},
\]
and the zero set of a nonzero polynomial has empty interior and
Lebesgue measure zero, the remaining conclusions follow.
\end{proof}

\paragraph{From local interpolation to the orbit scaffold.}
The selected depths in Equation~(\ref{eq:exact int eq}) provide one target-matching orbit position
for each anchor. To construct the dictionary, we retain the other
orbit positions as well, using
\[
J_kW^m x_{k,j},\qquad 0\leq m\leq d-1,
\]
as candidate atoms for sparse synthesis. Lemma \ref{lem:form orbits of W} shows that deeper
iterates contribute no additional directions, since they are scalar
multiples of these finite-depth iterates. Thus orbit interpolation
establishes the anchor constraints, while the full orbit family
provides the dictionary in which subsequent realizations are built.

When the original anchors also admit linear interpolants, these
provide local reference maps. The local-chart analysis in
Section \ref{subsect:scaffold selection} below quantifies how orbit interpolation changes the resulting
dictionary geometry and its bounded-coefficient sparse approximation
properties.

 Therefore, we build the scaffold from the full finite orbit family generated by these anchors. Specifically, define the orbit-based dictionary
\begin{equation}
\label{eq:def scaffold}
\mathcal D
:=
\Big\{
J_kW^m x_{k,j}
:\,
(x_{k,j},y_{k,j})\in D_k,\ 
1\leq k\leq K,\ 
0\leq m\leq d-1
\Big\}
\subseteq\mathbb{R}^C.
\end{equation}

We use this notation throughout:
\[
\mathcal I
:=\{(k,j,m):1\leq k\leq K,\ 1\leq j\leq s_k,\ 0\leq m\leq d-1\}, \qquad P:=|\mathcal I|.
\]
 We call $U:=\operatorname{span}\mathcal D$ the orbit scaffold and denote
$
\Pi_U:\mathbb{R}^C\to\mathbb{R}^C
$ the associated orthogonal projector
and write
\begin{equation}
\label{eq:def G}
    g_{k,j,m}:=J_kW^m x_{k,j}\in\mathbb R^C,  \qquad G:=[g_r]_{r\in\mathcal I}\in\mathbb R^{C\times P}.
\end{equation}

\[
\mathcal{I}_+:=\{r\in\mathcal{I}:\|g_r\|>0\},
\qquad q_r:=\frac{g_r}{\|g_r\|},\quad r\in\mathcal{I}_+.
\]

Define the fixed synthesis map associated with the orbit dictionary as
\begin{equation}
\label{eq:synthesis map def}
    S_{\mathcal D}:\mathbb R^P\longrightarrow U,
    \qquad
    S_{\mathcal D}(a)
    :=
    \sum_{(k,j,m)\in\mathcal I}a_{k,j,m}g_{k,j,m}=Ga,\qquad U=\operatorname{range}G.
\end{equation}

By Lemma~\ref{lem:form orbits of W}, depths $0\leq m\leq d-1$ contain all distinct orbit directions generated by $W$ up to scalar multiplication, since every deeper iterate is a scalar multiple of one of these residue classes.
 Moreover, the selected depths $L_{k,j}$ guarantee exact interpolation on the anchors. Thus, $\{y:(x, y)\in \Xi_{\rm sub}\}\subseteq \mathcal D$. So every column \(y_{k,j}\) of \(Y_k\) belongs to \(\mathcal D\). Conversely, since \(J_k=Y_kZ_k^\dagger\), every atom \(J_kW^m x_{k,j}\) belongs to \(\operatorname{range}(Y_k)\). Hence
\begin{equation}
\label{eq:U equal range Y}
U=\operatorname{span}\mathcal D
=\sum_{k=1}^K\operatorname{range}(Y_k)
=\operatorname{span}\{y:(x,y)\in\Xi_{\mathrm{sub}}\}.
\end{equation}
Therefore, for fixed anchors, changing the shift or partition changes the dictionary’s internal geometry, but not its full span.

\paragraph{The two realizability bottlenecks}
Let
$
f_\Xi:\mathbb{R}^d\to\mathbb{R}^C
$
be an empirical target map, and let
$
\hat f:\mathbb{R}^d\to\mathbb{R}^C.
$
Inserting the projected target and applying the triangle inequality gives
\begin{equation}
\label{eq:decomposition}
\|f_\Xi-\hat f\|_{\Xi_{\rm train}}
\leq
\underbrace{\|f_\Xi-\Pi_U f_\Xi\|_{\Xi_{\rm train}}}_{\varepsilon_{\mathrm{proj}}}
+
\underbrace{\|\Pi_U f_\Xi-\hat f\|_{\Xi_{\rm train}}}_{\varepsilon_{\mathrm{param}}}.
\end{equation}

The decomposition identifies the two structural bottlenecks governing realization on an orbit scaffold. The projection error
$
\varepsilon_{\mathrm{proj}}
$
measures how well $U$ captures the target set
$
\{f_\Xi x:(x,y)\in\Xi_{\rm train}\},
$
and depends only on the target map and the selected scaffold. The realization error
$
\varepsilon_{\mathrm{param}}
$
measures how accurately the projected target $\Pi_U f_\Xi$ is realized once $U$ has been fixed.
The remainder of this section addresses these two bottlenecks in order: first the selection of an orbit scaffold with small projection error, and then the constructive realization of the projected target on that scaffold.


\subsection{Scaffold selection and dictionary geometry}
\label{subsect:scaffold selection}
The existence of optimal scaffolds under which projection error collapses, is a mathematically well-posed problem. Indeed, by the scaffold identity (\ref{eq:U equal range Y}), each admissible anchor subset determines its
full scaffold through the span of its targets. Since the training set
is finite, the family of such scaffolds is finite, and the projection
error attains a minimum over any nonempty admissible family. Hence,  there exists $U_\star\in\mathcal U$ such that
$
\|f_\Xi-\Pi_{U_\star}f_\Xi\|_{\Xi_{\rm train}}
=
\min_{U\in\mathcal U}\|f_\Xi-\Pi_U f_\Xi\|_{\Xi_{\rm train}},
$
where $\mathcal U$ the set of all scaffolds relative to $\Xi_{\rm train}$. This guarantee is existential. We provide the details on the practical implementation of scaffold selection in Subsection~\ref{subsec:our framework in practice} below. We now derive an observable quantity serving as a pre-selection criterion.

\paragraph{Observable stability of orbit-sparse realizability}

By (\ref{eq:U equal range Y}), the full orbit scaffold is determined by the anchor targets, while the geometry of its generating dictionary depends on the chosen orbit configuration. We now show that the observable quantity $S(\mathcal P)$ defined in (\ref{eq:sp defintion}) below controls this internal geometry in a sense directly relevant to sparse realization.

Let $\mathcal P=\{D_1,\ldots,D_K\}$ be an orbit-admissible partition of $\Xi_{\mathrm{sub}}$, with
\[
D_k=\{(x_{k,j},y_{k,j})\}_{j=1}^{s_k},
\qquad
N=\sum_{k=1}^{K}s_k.
\]
Set \(X_k=[x_{k,1}\ \cdots\ x_{k,s_k}]\), and retain the preceding notation $Y_k,Z_k$ and $J_k=Y_kZ_k^\dagger$. Assume
\[
\ker X_k\subseteq\ker Y_k,
\qquad 1\leq k\leq K.
\]
Set $E_k:=\operatorname{range} Z_k$. The quantity
\begin{equation}
\label{eq:sp defintion}
S(\mathcal P):=
\frac{
\displaystyle\sum_{k=1}^{K}\sum_{j=1}^{s_k}
\frac{\|Y_k\|_F}{\sigma_{\min}^{+}(X_k)}
\left(
\operatorname{dist}(x_{k,j},E_k)
+
\frac{\|x_{k,j}\|}{\sigma_{\min}(Z_k)}
\left(
\sum_{\ell=1}^{s_k}
\|x_{k,\ell}-z_{k,\ell}\|^2
\right)^{1/2}
\right)
}{N}
\end{equation}
is therefore computable from the selected anchor configuration.

To identify what this quantity controls, define the canonical local-chart operator
\[
A_k:=Y_kX_k^\dagger.
\]
The compatibility assumption ensures that $A_kx_{k,j}=y_{k,j}$. Thus $A_k$ interpolates the original anchors, whereas $J_k$ interpolates their orbit representatives. This gives a natural reference dictionary
\[
\mathcal D_{\mathrm{chart}}
:=
\left\{
A_k\Pi_{E_k}W^m x_{k,j}:
(k,j,m)\in\mathcal I
\right\},
\]
whose atoms are generated by the original local charts restricted to the corresponding orbit-representative spaces. Comparing $\mathcal D_{\mathrm{chart}}$ with the actual orbit dictionary $\mathcal D$ isolates the geometric distortion introduced by orbit interpolation.

For any dictionary $\mathcal B$ with index set $\mathcal I$, let $S_{\mathcal B}$ denote its synthesis map. Given an integer $1\leq s\leq P$ and $B>0$, define the constrained sparse realization error
\[
e_{s,B}(y;\mathcal B)
:=
\inf_{\substack{
a\in\mathbb R^P\\
|\{r\in\mathcal I:a_r\neq0\}|\leq s\\
\sum_{r\in\mathcal I}|a_r|\leq B
}}
\|y-S_{\mathcal B}(a)\|.
\]
The coefficient bound prevents small atom perturbations from being arbitrarily amplified by large coefficients.
Finally, define the observable orbit-amplification factor
\[
\kappa(\mathcal P)
:=
\max_{(k,j,m)\in\mathcal I}
\frac{\|W^m x_{k,j}\|}
{\sum_{\ell=1}^{s_k}\|x_{k,\ell}\|}.
\]

\begin{theorem}[Observable stability of orbit-sparse realizability]
\label{thm:observable_sparse_stability}
Under the preceding assumptions, for every $y\in\mathbb R^C$, $1\leq s\leq P$, and $B>0$,
\[
\left|
e_{s,B}(y;\mathcal D)
-
e_{s,B}(y;\mathcal D_{\mathrm{chart}})
\right|
\leq
BN\kappa(\mathcal P)S(\mathcal P).
\]
\end{theorem}

\begin{proof}[Proof sketch]
The compatibility condition gives $Y_k=A_kX_k$, and hence
\[
J_k-A_k\Pi_{E_k}
=
A_k(X_k-Z_k)Z_k^\dagger.
\]
The resulting operator bound, combined with the definition of $S(\mathcal P)$, yields
\[
\max_{(k,j,m)\in\mathcal I}
\|(J_k-A_k\Pi_{E_k})W^m x_{k,j}\|
\leq
N\kappa(\mathcal P)S(\mathcal P).
\]
Thus synthesis with any common admissible coefficient vector differs between the two dictionaries by at most $BN\kappa(\mathcal P)S(\mathcal P)$. Taking the infimum over the same coefficient set proves the assertion. The details are given in Subsection~\ref{app:observable_sparse_stability}.
\end{proof}

Theorem \ref{thm:observable_sparse_stability} shows that $S(\mathcal P)$ controls the distortion of the entire bounded-coefficient sparse approximation problem, uniformly over targets and sparsity budgets. Consequently, among candidate anchor configurations with comparable orbit amplification, minimizing $S(\mathcal P)$ favors dictionaries whose sparse realization geometry remains close to that induced by the original local charts. This provides a structural justification for its use as a pre-selection criterion, independently of the projection quality of the full scaffold.




\subsection{Collapsing the parametrization error: the CCDS mechanism}
\label{subsect:realization error}

Once $U$ is fixed, standard sparse approximation methods such as Orthogonal Matching Pursuit (OMP) can be used to reduce the parametrization error, and in finite dimension can recover the orthogonal projection once a sufficient independent atom set has been selected. Our goal is therefore not to replace OMP merely as an approximation solver. Instead, we introduce a realization mechanism in which the residual trajectory itself becomes a controllable geometric object: Constructive Controlled Deviation Synthesis (CCDS).

\paragraph{Why CCDS if OMP already realizes the scaffold?}
OMP selects an atom, enlarges the current span, and refits by least squares; the resulting residual is therefore determined by the extremal orthogonal projection onto that span. 
CCDS retains the same span-building logic but replaces the rigid least-squares update by a controlled realization path: rather than accepting the residual dictated by the orthogonal projection at each step, it allows the residual decay to be prescribed within the geometry permitted by the current realization space. Thus, unlike OMP, which treats sparse realization as an endpoint approximation problem, CCDS turns it into a stepwise controllable process. 
 Both OMP and CCDS can accurately realize targets on a fixed scaffold, but OMP is confined to the extremal least-squares update on each selected span. CCDS can instead move through the interior of the feasible deviation geometry by prescribing admissible residual levels, while OMP has no parameter with which to realize those prescribed interior paths. This distinction is explicit in difficult geometry (see Experiments Section~\ref{sec:experiments}): at the minimal-deviation boundary CCDS agrees with OMP, whereas strict interior schedules generate feasible realization trajectories unavailable to OMP; under a fixed step budget, increasingly interior schedules deliberately trade residual contraction for greater path control. Thus deviation control enlarges the realizable trajectory class rather than uniformly improving endpoint approximation.

\paragraph{From Bernstein-type prescribed approximation to deviation control.}
The CCDS viewpoint is motivated by the Bernstein-type lethargy theory question of prescribing approximation errors along a nested family of subspaces, rather than merely taking the best approximation available at each stage. Let \(H\) be a real Hilbert space
\[
Y_0\subset Y_1\subset\cdots\subset Y_T\subset H,
\qquad
Y_t=Y_{t-1}\oplus\operatorname{span}\{u_t\},
\qquad
u_t\in Y_{t-1}^{\perp},\ \|u_t\|=1.
\]
Given a target $y\in H$, initialize $\widehat y_0=0$ and $r_0=y$, and update
\begin{equation}
\label{eq:update ccds}
\widehat y_t=\widehat y_{t-1}+\lambda_tu_t,
\qquad
r_t=r_{t-1}-\lambda_tu_t.
\end{equation}
Instead of forcing the extremal projection at every step, we prescribe a deviation
$
d_t:=\operatorname{dist}(r_t,Y_{t-1}).
$
Thus the entire stepwise realization problem reduces to the scalar distance equation
\begin{equation}
\label{eq:distance eq}
\operatorname{dist}(r_{t-1}-\lambda_tu_t,Y_{t-1})=d_t.
\end{equation}
The next result completely characterizes when this equation is feasible and gives the corresponding coefficient $\lambda_t$ explicitly, thereby turning prescribed approximation geometry into a constructive finite-horizon realization mechanism. 

\begin{theorem}[Constructive controlled-deviation principle]
\label{thm:controlled deviation principle}
Under the preceding notation, for each $t$, define
\[
a_t:=(I-\Pi_{Y_{t-1}})r_{t-1},
\qquad
\alpha_t:=\langle a_t,u_t\rangle,
\qquad
d_{t,\min}:=\sqrt{\|a_t\|^2-\alpha_t^2}.
\]
Then the prescribed deviation equation (\ref{eq:distance eq})
is feasible if and only if $d_t\ge d_{t,\min}$, in which case
\begin{equation}
\label{eq:lambda t expression}
\lambda_t
=
\alpha_t\pm\sqrt{d_t^2-d_{t,\min}^2}.
\end{equation}
Consequently, every feasible deviation profile
$(d_t)_{t=1}^T$ can be realized constructively by the updates (\ref{eq:update ccds})
satisfying exactly
$
\operatorname{dist}(r_t,Y_{t-1})=d_t,
$ for all $ t=1,\ldots,T.
$
\end{theorem}

\begin{proof}
We proceed sequentially in $t$.
Suppose that $\lambda_1,\ldots,\lambda_{t-1}$ have already been chosen,
so that $\hat y_{t-1}\in Y_{t-1}$ and
$r_{t-1}=y-\hat y_{t-1}$ are determined. Since
$u_t\in Y_t\cap Y_{t-1}^{\perp}$ and $\|u_t\|=1$.
By Lemma~\ref{lem:distance-equation}, the equation
$
\operatorname{dist}(r_{t-1}-\lambda u_t,Y_{t-1})=d_t
$
admits a real solution if and only if
$
d_t\ge
\sqrt{\|a_t\|^2-\alpha_t^2}
=
d_{t,\min}.
$
Under the assumed feasibility condition, its solutions are precisely
\[
\lambda
=
\alpha_t\pm\sqrt{d_t^2-d_{t,\min}^2}.
\]
Choose one of these two roots according to the prescribed deterministic
rule and denote it by $\lambda_t$. Define
\[
\hat y_t:=\hat y_{t-1}+\lambda_tu_t,
\qquad
r_t:=y-\hat y_t.
\]
Since $\hat y_{t-1}\in Y_{t-1}\subset Y_t$ and $u_t\in Y_t$, one has
$\hat y_t\in Y_t$. Moreover,
\[
r_t
=
y-\hat y_{t-1}-\lambda_tu_t
=
r_{t-1}-\lambda_tu_t.
\]
Therefore, by the defining choice of $\lambda_t$,
\[
\operatorname{dist}(r_t,Y_{t-1})
=
\operatorname{dist}(r_{t-1}-\lambda_tu_t,Y_{t-1})
=
d_t.
\]

Starting from $\hat y_0=0$ and $r_0=y$, this construction can be applied
successively for $t=1,\ldots,T$. Hence every feasible profile
$(d_t)_{t=1}^T$ is realized, and the resulting coefficients satisfy
Equation (\ref{eq:lambda t expression}).
The deterministic root-selection rule makes the resulting sequence
$(\lambda_t)_{t=1}^T$ deterministic.
\end{proof}

\begin{remark}[Prescribed deviation versus total residual]
\label{rem:dt vs rt}
For nested orthogonal innovations, under the recursion
\(r_t=r_{t-1}-\lambda_tu_t\), we have, for every \(t\geq1\),
\begin{equation}
\label{eq:prescribed-distance}
\|r_t\|^2
=
d_t^2+
\sum_{s=1}^{t-1}
\bigl(d_s^2-d_{s,\min}^2\bigr).
\end{equation}
Equation~\eqref{eq:prescribed-distance} shows that prescribing
\(d_s=d_{s,\min}\) at every step eliminates the accumulated
coordinate-error terms and reduces the total residual exactly to \(
d_{t,\min}
=\operatorname{dist}(y,Y_t)\). This helps explain the very
small parametrization errors observed in the experiments. Interior
schedules retain a controlled excess residual in exchange for pathwise
deviation control. A complete derivation
of Equation~\eqref{eq:prescribed-distance} is given in
Subsection~\ref{app:CCDS proofs}.
\end{remark}

We next instantiate this abstract deviation-control principle on the orbit dictionary used by our realization framework.

\paragraph{Dictionary-induced CCDS realization}
To instantiate Theorem~\ref{thm:controlled deviation principle} on the
orbit dictionary, let $\mathcal D\subset H$ be unit norm. At step $t$,
given $Y_{t-1}$ and $r_{t-1}$, select an atom
$q_t\in\mathcal D\setminus Y_{t-1}$ and define
\begin{equation}
\label{eq:gram-schmidt}
\eta_t:=\|q_t-\Pi_{Y_{t-1}}q_t\|,\qquad
u_t:=\frac{q_t-\Pi_{Y_{t-1}}q_t}{\eta_t},
\qquad
Y_t:=Y_{t-1}\oplus\operatorname{span}\{u_t\}.
\end{equation}
For any feasible prescribed deviation $d_t\ge d_{t,\min}$,
Theorem~\ref{thm:controlled deviation principle} determines
$\lambda_t$ explicitly and updates
\[
\widehat y_t=\widehat y_{t-1}+\lambda_tu_t,
\qquad
r_t=y-\widehat y_t.
\]

Since $q_t\notin Y_{t-1}$ and
$Y_t=\operatorname{span}\{q_1,\ldots,q_t\}$, the selected atoms
$q_1,\ldots,q_T$ are linearly independent. Moreover,
$u_t\in\operatorname{span}\{q_1,\ldots,q_t\}$ for every $t$.
Consequently, for every CCDS path $q_{1:T}$ there exists a unique
$\beta^{(T)}\in\mathbb R^T$ such that
\[
\widehat y_T
=
\sum_{t=1}^T\lambda_tu_t
=
\sum_{j=1}^T\beta_j^{(T)}q_j.
\]
We call $\beta^{(T)}$ the \emph{CCDS dictionary-coordinate vector}.
Its explicit triangular recursion in terms of the
innovation coordinates is given in Subsection~\ref{app:CCDS proofs}.
 We denote this path-specific realization by
\[
\operatorname{CCDS}_{\mathcal D}^{T}
(y;q_{1:T},d_{1:T})
:=
\widehat y_T,
\]
for the CCDS output after $T$ steps. Here $q_{1:T}:=(q_1,\ldots,q_T)$ denotes the prescribed atom ordering and
$d_{1:T}:=(d_1,\ldots,d_T)$ the prescribed deviation profile.

For a target $y$, let $T(y)$ denote the termination horizon of its CCDS path. We denote by
\[
    \beta_{\mathcal D}(y)\in\mathbb R^P
\]
the full-dictionary CCDS coordinate vector obtained by placing the path-specific coefficients $\beta^{(T(y))}$ in the selected dictionary coordinates and setting every unselected coordinate equal to zero. If CCDS is run using normalized atoms, the corresponding normalization factors are absorbed into $\beta_{\mathcal D}(y)$ so that synthesis is always performed with the original atoms $d_{k,j,m}$. We fix the atom-selection rule, deviation profile, root choices, and all tie-breaking conventions; consequently, $y\mapsto\beta_{\mathcal D}(y)$ is a well-defined map. With this notation,
\begin{equation}
    \operatorname{CCDS}_{\mathcal D}(y)
    =S_{\mathcal D}\bigl(\beta_{\mathcal D}(y)\bigr).
    \label{eq:ccds-full-coordinate}
\end{equation}

The nested orthogonal-innovation construction is designed so that, at the minimal deviation boundary, CCDS recovers the Fourier coefficients of the target in the induced orthonormal basis. Proof of Proposition \ref{prop:minimal dev ccds recovers ortho coordinates} in Subsection \ref{app:CCDS proofs}.

\begin{proposition}[Minimal-deviation CCDS recovers the orthogonal coordinates]
\label{prop:minimal dev ccds recovers ortho coordinates}
Let $\mathcal D\subset H$ be a unit-norm dictionary, let
$q_1,\ldots,q_s\in\mathcal D$ be linearly independent, and set
$
E:=\operatorname{span}\{q_1,\ldots,q_s\}.
$
For $y\in H$, apply CCDS with atom ordering $q_{1:s}$ and
$d_t=d_{t,\min}$ for $t=1,\ldots,s$. Then
\[
\operatorname{CCDS}_{\mathcal D}^{s}
\bigl(y;q_{1:s},d_{\min,1:s}\bigr)
=
\Pi_E y=\sum_{t=1}^s\lambda_tu_t
 =\sum_{j=1}^s\beta_j^{(s)}q_j,
\]
where $\lambda_t=\langle y,u_t\rangle$ and $\beta^{(s)}$ is the CCDS dictionary-coordinate vector.
\end{proposition}

\paragraph{MP and OMP as minimal-deviation boundaries.} Neither  Matching Pursuit (MP) nor OMP is, in general, a special case of the other. CCDS, however, provides a common geometric mechanism containing both as boundary regimes. The following proposition identifies MP and OMP as the two minimal-deviation boundary regimes of CCDS while strict interior CCDS updates are globally inaccessible to OMP, marking a geometric separation between OMP and CCDS. Complete proof in Subsection \ref{app:CCDS proofs}.
\begin{proposition}
\label{prop:MP OMP boundaries CCDS}
Let $\mathcal D\subset H$ be a unit-norm dictionary.

\emph{(i) (MP boundary).}
If $Y_{t-1}=\{0\}$ at every step, $q_t$ is selected by the MP
atom-selection rule, and $d_t=d_{t,\min}$, then CCDS coincides with
MP.

\emph{(ii) (OMP boundary).}
If $q_t$ is selected by the OMP atom-selection rule such that the nested spaces
$
Y_t=\operatorname{span}\{q_1,\ldots,q_t\}
$, and
$d_t=d_{t,\min}$, then CCDS coincides with OMP
until termination.

\emph{(iii) (Global geometric separation from OMP).}
Any interior CCDS update with
$
d_t>d_{t,\min},
\
\lambda_t\neq0,
$
departing from an OMP boundary state
$
\widehat y_{t-1}=\Pi_{Y_{t-1}}y
$
is globally inaccessible to OMP, irrespective of its atom selection or ordering.
\end{proposition}

\begin{proof}[Proof sketch]
At minimal deviation, \(\lambda_t=\alpha_t\), giving the MP update
with a trivial reference space and the OMP projection update with
nested selected spaces. Every OMP iterate satisfies
\[
\langle y-\widehat y_{\mathrm{OMP}},
        \widehat y_{\mathrm{OMP}}\rangle=0.
\]
For a CCDS update departing from
\(\widehat y_{t-1}=\Pi_{Y_{t-1}}y\), orthogonality gives
\[
\langle y-\widehat y_t,\widehat y_t\rangle
=\lambda_t(\alpha_t-\lambda_t).
\]
Strict interior deviation implies \(\lambda_t\neq\alpha_t\).
If also \(\lambda_t\neq0\), this violates the invariant of every
OMP iterate, irrespective of its selected subspace.
\end{proof}

\paragraph{Beyond the MP/OMP boundary.}
The additional freedom of CCDS appears when
$d_t>d_{t,\min}$. In this regime,
the update is no longer forced to be the extremal orthogonal
projection. Instead, CCDS can prescribe any feasible deviation profile
$(d_t)_{t=1}^T$ and thereby control the realization path through the
nested spaces $Y_t$.

The strict regime $d_t>d_{t,\min}$ also gives CCDS a pathwise stability
property. The perturbation stability Theorem \ref{thm:stability control interpolation} below shows that, when the prescribed
CCDS path has a positive feasibility margin, the same atom labels,
deviations, and root branches remain feasible under sufficiently small
perturbations of the orbit dictionary, and the transported realization
varies locally Lipschitz continuously. The boundary case $d_t=d_{t,\min}$, which recovers OMP for the same atom-selection rule, is not covered by this argument because the feasibility margin vanishes. Thus the distinction from OMP is not in endpoint realizability, but the ability to prescribe a realization path and obtain a stability guarantee for that prescribed path.
At the coefficient level, interior choices $d_t>d_{t,\min}$ move CCDS away from the boundary where the root term
vanishes in
$
\lambda_t=\alpha_t\pm\sqrt{d_t^2-d_{t,\min}^2}.
$
This avoids boundary-root degeneracy.

\paragraph{Intrinsic orbit-certificate sparsity.}
\label{sub:CCDS sparsity}

The deviation-control mechanism is compatible with the intrinsic sparse
geometry of the orbit dictionary. Let
$\mathcal D$ be the orbit dictionary, $U=\operatorname{span}\mathcal D$,
and
$
\mathcal I
$ the dictionary index set.
Write $q_{k,j,m}\in\mathcal D$ for the corresponding unit-norm
orbit atom. For $S\subseteq\mathcal I$, define
\[
E_S:=\operatorname{span}\{q_{k,j,m}:(k,j,m)\in S\}.
\]
For $(x_i,y_i)\in\Xi_{\mathrm{train}}$ and
$0\leq\eta\leq\|\Pi_Uy_i\|$, set
\[
s_i(\eta)
:=
\min\Bigl\{
|S|:
S\subseteq\mathcal I,\ 
\operatorname{dist}(\Pi_Uy_i,E_S)\leq\eta
\Bigr\}.
\]
Thus $s_i(\eta)$ measures the intrinsic per-sample number of orbit-dictionary atoms
needed to realize the projected target to tolerance $\eta$.
Let
$
S_i=\{r_1,\ldots,r_s\}
$
be a minimizer, so that $s=|S_i|=s_i(\eta)$. The atoms
$q_{r_1},\ldots,q_{r_s}$ are linearly independent. Indeed, if one
belonged to the span of the others, removing its index would leave
$E_{S_i}$ unchanged, contradicting the minimality of $S_i$.
Proposition~\ref{prop:minimal dev ccds recovers ortho coordinates}
therefore applies directly to this atom family. Since $E_{S_i}\subseteq U$,
we have $
\Pi_{E_{S_i}}y_i
=
\Pi_{E_{S_i}}\Pi_Uy_i.
$
Hence, by the defining property of the minimizing support $S_i$,
\[
\left\|
\Pi_Uy_i-\Pi_{E_{S_i}}y_i
\right\|
=
\operatorname{dist}(\Pi_Uy_i,E_{S_i})
\leq\eta.
\]
Thus, we have the following.

\begin{corollary}[Intrinsic local orbit-certificate sparsity of full CCDS]
\label{thm:CCDS sparsity certificate}
With the notation above,
\[
\operatorname{CCDS}_{\mathcal D}^{s}
\bigl(y_i;q_{r_1:r_s},d_{\min,1:s}\bigr)
=
\Pi_{E_{S_i}}y_i
=
\sum_{j=1}^s\beta_j^{(s)}q_{r_j}, 
\]
and
\[
\left\|
\Pi_Uy_i-
\operatorname{CCDS}_{\mathcal D}^{s}
\bigl(y_i;q_{r_1:r_s},d_{\min,1:s}\bigr)
\right\|
\leq\eta.
\]
In particular, this CCDS realization is supported on at most
$s_i(\eta)$ orbit-dictionary atoms.
\end{corollary}

\paragraph{Repairing CCDS infeasibility.}
Prescribing the realization path introduces a feasibility obstruction with no direct analogue in standard OMP: for a chosen innovation direction $u_t$, a desired deviation
$d_t<d_{t,\min}$ is infeasible in the current ambient space. This is the
finite-dimensional analogue of the prescribed-approximation obstruction
underlying Bernstein-type lethargy theory. Such infeasibility can be
repaired constructively: by adjoining one orthogonal dimension, any
prescribed $0\le d_t\le\|a_t\|$ can be made feasible through a suitable
lifted innovation direction. The explicit lifting theorem and its proof
are given in Subsection~\ref{subsec:ccds feasibility obstruction and lifting}.

\subsection{Stability under relaxed block interpolation}
Exact interpolation provides the canonical construction, with block maps $J_k=Y_kZ_k^\dagger$ recovering the anchor outputs exactly. In practice, however, we use the regularized interpolant
$
J_{k,\lambda}:=Y_k(Z_k^\top Z_k+\lambda I)^{-1}Z_k^\top,
\
\lambda>0,
$
which is better conditioned on noisy or nearly singular blocks. The next theorem shows that this relaxation preserves the scaffold geometry and perturbs the CCDS realization error only continuously, thereby connecting the exact theory to the implementation used in practice. Complete proof in Subsection \ref{subsec:stability theorems}.

\begin{theorem}[Stability of the scaffold/CCDS bottlenecks under relaxed interpolation]
\label{thm:stability control interpolation}
Fix the orbit data
$
(W,\Xi_{\rm sub},\{D_k\}_{k=1}^K).
$
For each block write \(Z_k\) for the orbit matrix and \(Y_k\) for the corresponding output matrix as defined in (\ref{eq:matrices Z and Y}), let
\(J_k=Y_kZ_k^\dagger\) be the reference minimal-norm exact interpolant. Let \(G\) be the associated atom matrix as in (\ref{eq:def G}) such that
$
U=\operatorname{range}(G),\ r:=\operatorname{rank}(G),\ \gamma:=\sigma_r(G)>0.
$
Assume the reference CCDS realizer $\rm CCDS_{\mathcal D}$ on \(U\) has a finite prescribed path with positive feasibility margin and selected raw directions uniformly bounded away from zero.
 For every admissible relaxed block family
$
J_k'=Y_kR_k',
$
where \(R_k'\) is a regularized pseudoinverse factor for the same block data, \(G'\) the corresponding atom matrix and \(U':=\operatorname{range}(G')\), there exist \(\delta_0>0\) such that if
$
\|G'-G\|_2<\delta_0,
$
then \(U'=U\). Consequently,
\[
\varepsilon_{\rm proj}(f_\Xi,U')
=
\varepsilon_{\rm proj}(f_\Xi,U).
\]
Moreover, transporting the reference CCDS path to the perturbed atoms, with the same prescribed deviations and the same root branches, gives a CCDS realizer $\rm CCDS_{\mathcal D'}$ on \(U'\) satisfying
\[
\varepsilon_{\rm param}(f_\Xi,U', \rm CCDS_{\mathcal D'})
\le
\varepsilon_{\rm param}(f_\Xi,U, \rm CCDS_{\mathcal D})
+
C_{\rm CCDS}\|G'-G\|_2.
\]
Thus, the projection bottleneck is invariant under relaxed interpolation and the constructive CCDS bottleneck is locally Lipschitz stable. 
\end{theorem}

\begin{proof}[Proof sketch]
Set \(\eta=\|G'-G\|_2\). Since \(J_k'=Y_kR_k'\), the scaffold
identity gives \(U'\subseteq U\). For \(\eta<\gamma/2\), Weyl's
inequality yields
\[
\sigma_r(G')\geq\gamma-\eta>0,
\]
so \(U'=U\) and the projection errors coincide.

Transport the same atom labels, deviations, and root branches.
The positive innovation norms preserve the normalization steps,
while the positive feasibility margins keep the square-root
arguments away from zero. Consequently, the innovation directions
and coefficient updates vary locally Lipschitz continuously.
Induction along each finite path gives
\[
\|r_t'-r_t\|\leq C_t\eta.
\]
Taking a common constant over the finitely many training paths and
averaging the terminal residual bounds proves the stated
parametrization-error estimate.
\end{proof}


\subsection{Our framework in practice.}
\label{subsec:our framework in practice}

The preceding theory isolates the projection and parametrization bottlenecks and provides constructive mechanisms for controlling them. In practice, however, scaffold construction begins one step earlier: from the anchor subset $\Xi_{\mathrm{sub}}$, one must choose a partition, orbit depths, and a wrapped shift $W$ that generate a suitable candidate orbit geometry. In Subsection \ref{subsect:scaffold selection} we derived observable quantities such as $S(\mathcal P)$ guiding this preliminary selection. Now we describe another criterion and then summarize how they enter the two realizability phases.

\paragraph{Vanishing-cycle criterion.}
A second, independent criterion concerns the repeated orbit dynamics of $W$. Set
$
\rho:=\prod_{i=0}^{d-1}w_i.
$
If $|\rho|<1$, then repeated full cycles decay geometrically. Indeed, if $m=qd+r$ with $0\leq r<d$, Lemma~\ref{lem:form orbits of W} gives
$
W^m=\rho^qW^r
$
and hence
\begin{equation}
\label{eq:noise supp orbit decay}
\|W^m z\|
\leq
|\rho|^{\lfloor m/d\rfloor}C_z,
\qquad
C_z:=\max_{0\leq r\leq d-1}\|W^r z\|.
\end{equation}
Thus, if an input perturbation $\eta$ is propagated through the orbit, $\|W^m\eta\|\leq |\rho|^{\lfloor m/d\rfloor}C_\eta$; hence  $|\rho|<1$ acts as an implicit noise regularizer by geometrically damping perturbations propagated through repeated orbit cycles, without changing the formal scaffold generated by the $d$ residue-class depths. This is the main reason we favor vanishing shifts in scaffold discovery.

 The practical pipeline retains the same
structural decomposition developed above. Phase~1 addresses the projection
bottleneck by constructing and selecting an orbit scaffold, Phase~2 addresses
the parametrization bottleneck by constructively realizing the projected
targets on the selected scaffold, and Phase~3 extends the resulting CCDS
coefficient certificates to unseen inputs. Let
\(\Xi_{\mathrm{sub}}\subseteq\Xi_{\mathrm{train}}\), where
\(\Xi_{\mathrm{sub}}\) is the anchor subset used to construct the orbit
scaffold.

\paragraph{Phase~1: scaffold discovery and selection.}
Phase 1 searches over a finite collection of candidate anchor configurations
\[
\vartheta=(\Xi_{\mathrm{sub}},\mathcal{P},W),
\]
where \(\Xi_{\mathrm{sub}}\) is an anchor subset, \(\mathcal{P}\) is a partition of the anchors, and \(W\) is a wrapped weighted shift. For each candidate, orbit-independence tests select admissible depths \(L_{k,j}\), and the corresponding block interpolants are computed as
\[
J_k=Y_k Z_k^\dagger,
\qquad
Z_k=[W^{L_{k,1}}x_{k,1}\ \cdots\ W^{L_{k,s_k}}x_{k,s_k}].
\]
The candidate orbit dictionary and associated orbit scaffold are then
\begin{equation}
\label{eq:candidate orbit scaffold}
\mathcal{D}_{\vartheta}
=
\left\{
J_kW^m x_{k,j}:
1\leq k\leq K,\,
1\leq j\leq s_k,\,
0\leq m\leq d-1
\right\},
\qquad
U_{\vartheta}:=\operatorname{span}\mathcal{D}_{\vartheta}.
\end{equation}

The code first constructs several disjoint farthest-first anchor candidates. For each one, it forms a greedy orbit-admissible partition, refines the partition until the prescribed block-size condition is reached, computes the local-chart score \(\mathcal{S}(\mathcal{P})\), and constructs the corresponding block interpolants. A wrapped shift is then fitted or constrained in the vanishing-cycle regime, and the full candidate orbit dictionary \(\mathcal{D}_{\vartheta}\) is assembled.

To evaluate the projection quality of a candidate, the code computes an orthonormal SVD chain inside \(U_{\vartheta}=\operatorname{span}\mathcal{D}_{\vartheta}\) . This chain provides a basis for a nested sequence of projection spaces and a numerical diagnostic of the residual decay. It is not itself the dictionary used for synthesis. The candidate configuration is retained according to the smallest projection diagnostic among the evaluated candidates, with early termination when the prescribed tolerance is reached. The output of Phase~1 is therefore the selected configuration
\[
\vartheta^\star
=
(\Xi_{\mathrm{sub}}^\star,\mathcal{P}^\star,W^\star),
\]
together with the orbit dictionary \(\mathcal{D}_{\vartheta^\star}\) and its full span
\[
U_{\vartheta^\star}
=
\operatorname{span}\mathcal{D}_{\vartheta^\star}.
\]

\paragraph{Phase~2: CCDS realization.}
Once the scaffold has been selected, Phase~2 addresses the
parametrization bottleneck by realizing each projected training target
using the fixed orbit dictionary $\mathcal D_{\vartheta^\star}$ defined in
(\ref{eq:def scaffold}). 
For each $(x_i,y_i)\in\Xi_{\mathrm{train}}$, define its coefficient certificate and realized target by
\begin{equation}
    \beta_i:=\beta_{\mathcal D_{\vartheta^\star}}(y_i),
    \qquad
    c_i:=S_{\mathcal D_{\vartheta^\star}}(\beta_i)
    =\operatorname{CCDS}_{\mathcal D_{\vartheta^\star}}(y_i)\in U.
    \label{eq:training-certificates}
\end{equation}
Once $\mathcal D_{\vartheta^\star}$ and the CCDS rule have been fixed, the construction in~\eqref{eq:training-certificates} depends on the target $y_i$ alone. The input $x_i$ records the location in input space at which this certificate is observed. Thus Phase~2 produces the coefficient-labeled cloud
\[
    \mathcal C_{\mathrm{train}}
    :=\{(x_i,\beta_i,c_i)\}_{i=1}^n,
\]
and Phase~3 extends this finite collection to unseen inputs.

Phases 1 and 2 have now identified a scaffold, shown that its projected target is constructively and sparsely realizable on each training target, and established that this mechanism survives the relaxed interpolation used in practice.

\paragraph{Phase 3: Prediction from CCDS certificates.}
 Phase 3 extend Phases 1-2 realizations to unseen inputs. Phase 3 coefficient extension map is defined by
\[
\mathcal E_{\vartheta^\star}(\,\cdot\,;\mathcal C_{\rm train}):
\mathbb R^d\longrightarrow\mathbb R^P,
\qquad
\mathcal C_{\rm train}=\{(x_i,\beta_i,c_i)\}_{i=1}^n,
\]
fitted from the training data. The associated predictor is
\begin{equation}
\label{eq:prediction rule}
\widetilde f_{\vartheta^\star}(x)
=S_{\mathcal D_{\vartheta^\star}}\bigl(\mathcal E_{\vartheta^\star}(x;\mathcal C_{\rm train})\bigr).
\end{equation}

\begin{definition}[Weighted Backward Shift Neural Network]
\label{def:wbsnn}
A \emph{Weighted Backward Shift Neural Network} (WBSNN) associated with $\Xi_{\mathrm{train}}$ is the predictor $\widetilde f$ of (\ref{eq:prediction rule}) produced by the three-phase construction of this paper.
\end{definition}

In the predictive experiments, $\mathcal E_{\vartheta^\star}(x;\mathcal C_{\rm train})=\alpha_\theta(x)$ is a learned coefficient head. Writing $g_\theta=S_{\mathcal D}\circ\alpha_\theta$, it may be trained with
\[
\mathcal L_3(\theta)=\frac1n\sum_{i=1}^n
\left[
\ell_{\rm pred}\bigl(g_\theta(x_i),y_i\bigr)
+\lambda_{\rm CCDS}\|g_\theta(x_i)-c_i\|_2^2
+\lambda_{\rm coeff}\|\alpha_\theta(x_i)-\beta_i\|_2^2
+\lambda_1\|\alpha_\theta(x_i)\|_1
\right],
\]
where the nonnegative weights specify which forms of constructive supervision are used.

The distinction between realization and prediction is visible without imposing a particular extension rule. For any test pair $(x,y)$ and any training index $i$,
\[
\begin{aligned}
y-\widetilde f_{\vartheta^\star}(x)
={}&(y-\Pi_{U_{\vartheta^\star}} y)
+\Pi_{U_{\vartheta^\star}} (y-y_i)
+(\Pi_{U_{\vartheta^\star}} y_i-c_i)\\
&+(c_i-\widetilde f_{\vartheta^\star}(x_i))
+(\widetilde f_{\vartheta^\star}(x_i)-\widetilde f_{\vartheta^\star}(x)).
\end{aligned}
\]
The five terms expose scaffold projection, target transfer, CCDS realization, training prediction, and predictor variation.

\section{Experiments}
\label{sec:experiments}

We evaluate three separate questions: whether a reduced scaffold captures target energy, whether CCDS realizes a projected target with controlled deviations, and whether a learned extension predicts unseen inputs. We then examine dictionary structure, relaxed interpolation, and discovery budget. Reported uncertainties are sample standard deviations across the stated seeds. Configuration details and the recorded implementation conventions appear in Appendix~\ref{sec:experiment-config}.

The experiments are designed as mechanism tests rather than as an unconstrained benchmark contest. All matched baselines operate on the same compressed features and training information, while the orbit scaffold is constructed from only a small discovery subset. This separates the contributions of scaffold geometry, constructive realization, and coefficient prediction, and tests whether the proposed structure provides useful data and representation leverage under controlled computational conditions.

\subsection{Scaffold geometry and controlled realization}

\paragraph{Target energy at matched rank}
On a synthetic linear-response problem, we compare spectral selection within the candidate orbit span with random orbit-atom selection and Gaussian subspaces of the same rank. Across ten anchor seeds, each random reference averages twenty draws per rank. Table~\ref{tab:scaffold-rank} shows that two selected directions capture \(99.05\%\) of target energy, compared with \(34.74\%\) for random orbit selection and \(28.35\%\) for Gaussian subspaces. Across ranks one through six, the selected scaffold has lower projection error than the random orbit reference in \(57/60\) seed--rank comparisons and than the Gaussian reference in all sixty. This experiment evaluates the maximal-drop spectral endpoint of the scaffold-selection principle; it does not test an interior prescribed energy profile.

\begin{table}[t]
\centering
\small
\begin{tabular}{crrr}
\hline
Rank & Selected & Random orbit & Gaussian \\
\hline
1 & \(0.1586\pm0.0062\) & \(0.5798\pm0.0184\) & \(0.5929\pm0.0163\) \\
2 & \(0.0620\pm0.0122\) & \(0.5117\pm0.0247\) & \(0.5352\pm0.0252\) \\
6 & \(0.0175\pm0.0275\) & \(0.1490\pm0.0739\) & \(0.2041\pm0.0134\) \\
\hline
\end{tabular}
\caption{Mean Euclidean projection error at matched rank, averaged over ten anchor seeds. Spectral selection and random orbit selection are restricted to the same candidate orbit span; Gaussian subspaces are sampled in the ambient output space.}
\label{tab:scaffold-rank}
\end{table}

\begin{figure}[t]
\centering
\includegraphics[width=0.5\linewidth]{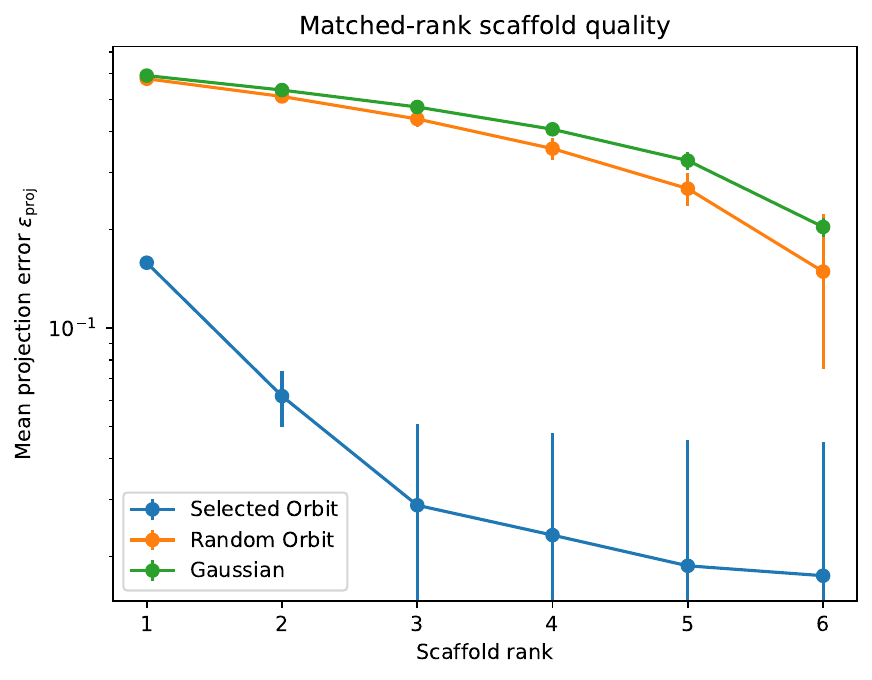}
\caption{Matched-rank scaffold quality. Mean projection error across scaffold ranks for the selected orbit, random orbit, and Gaussian constructions. Error bars show the reported standard deviations.}
\label{fig:matched-rank-scaffold-quality}
\end{figure}

Because the compared subspaces have matched rank, and the orbit alternatives are restricted to the same candidate span, this result isolates the effect of data-adaptive scaffold selection. The gain cannot be attributed merely to a larger subspace or a more favorable ambient dimension: the selected orbit directions capture the target geometry substantially more effectively than either random orbit choices or unstructured Gaussian directions. Figure~\ref{fig:matched-rank-scaffold-quality} shows the selected scaffold's lower mean projection error across all six ranks.

\paragraph{The OMP boundary and interior paths}
A weak-alignment stress test uses 500 inputs, a 350-atom dictionary, and a seven-step nested realization. At the minimal-deviation boundary, CCDS and OMP on the same dictionary have mean realization errors \(7.31\times10^{-8}\) and \(7.35\times10^{-8}\), respectively; their outputs differ by \(9.62\times10^{-8}\) on average. Interior schedules remain feasible for all 500 inputs, while exposing the distinction between prescribed deviation and terminal error (Table~\ref{tab:interior-paths}). With \(\gamma=0.5\), \(96.8\%\) of inputs encounter an interior step; with \(\gamma=0.9\), all do. The larger terminal errors reflect coordinate discrepancies retained from earlier innovations, as predicted by the residual identity in Remark \ref{rem:dt vs rt}. Thus interior control supplies additional path freedom, rather than an automatic improvement in endpoint accuracy.

This comparison establishes the intended separation between endpoint realization and path control. At the minimal-deviation boundary, CCDS reproduces the residual-minimizing behavior associated with the selected nested span. Interior schedules remain feasible while deliberately retaining controlled coordinate discrepancies, thereby providing a path-level degree of freedom unavailable at the boundary. In the nested orthogonal setting, OMP is confined to the minimal-deviation boundary: its least-squares refit selects \(\lambda_t=\alpha_t\) and therefore cannot realize a prescribed interior deviation \(d_t>d_{t,\min}\). Thus the interior CCDS trajectories are genuinely inaccessible to OMP on the same dictionary, rather than merely alternative OMP choices.

\begin{table}[t]
\centering
\small
\begin{tabular}{crrr}
\hline
\(\gamma\) & Mean \(\varepsilon_{\mathrm{param}}\) & Interior steps (\%) & Mean support \\
\hline
0 & \(7.31\times10^{-8}\) & 0 & 7.00 \\
0.5 & \(6.92\times10^{-2}\) & 34.77 & 7.00 \\
0.9 & \(5.11\times10^{-1}\) & 76.43 & 6.99 \\
\hline
\end{tabular}
\caption{Nested CCDS on the weak-alignment problem. The dictionary, target responses, and seven-step horizon are fixed; the interior schedule changes.}
\label{tab:interior-paths}
\end{table}

\paragraph{Interior realization across operator families}
We next hold a randomly constructed 900-atom dictionary fixed and evaluate eight linear operator families on 2,000 inputs each. A six-step CCDS schedule allocates a relative terminal error budget of \(10^{-3}\). All 16,000 paths complete without a reported numerical failure. Depending on the family, \(57.10\%\)--\(70.65\%\) of inputs encounter an interior step, with mean realization errors between \(1.51\times10^{-4}\) and \(7.71\times10^{-4}\) and mean supports between 5.99 and 6.00. The separate trivial-reference solver also realizes these targets, with errors between \(2.40\times10^{-8}\) and \(1.06\times10^{-7}\). These results demonstrate feasible sparse interior paths across different response geometries; the low frequency of the logged MP-branch certificate does not establish failure of MP or OMP.

The fact that the same sparse construction remains feasible across substantially different operator geometries is important: the observed behavior is not tied to one favorable response map. The experiment supports CCDS as a geometry-aware realization mechanism whose interior certificates remain usable even when the available dictionary does not provide a uniformly dominant greedy direction.

\subsection{Prediction from constructive certificates}

The main predictive runs use global coefficient heads on Gas Sensor Drift, ISOLET, and MNIST, with five runs per setting. The tabular baselines receive the same feature representation and training subset. Table~\ref{tab:predictive-results} reports held-out accuracy; the MNIST holdout is the final 12,000 examples of the official training pool, rather than the official test set.

\begin{table}[t]
\centering
\small
\setlength{\tabcolsep}{3pt}
\begin{tabular}{lrrrrrr}
\hline
Setting & WBSNN & LR & RF & RBF SVM & MLP & GBT \\
\hline
Gas, 25 & \(74.23\pm2.23\) & \(63.65\pm2.98\) & \(62.34\pm4.57\) & \(52.25\pm4.03\) & \(58.56\pm6.70\) & \(58.25\pm6.94\) \\
Gas, 128 & \(71.98\pm1.59\) & \(69.11\pm1.19\) & \(64.58\pm2.80\) & \(60.38\pm0.64\) & \(64.03\pm3.44\) & \(65.98\pm2.13\) \\
ISOLET, 25 & \(92.42\pm0.43\) & \(91.52\pm0.11\) & \(88.33\pm0.44\) & \(92.88\pm0.00\) & \(91.66\pm0.42\) & \(88.97\pm0.29\) \\
ISOLET, 100 & \(94.47\pm0.34\) & \(92.82\pm0.00\) & \(91.62\pm0.48\) & \(95.06\pm0.00\) & \(93.66\pm0.60\) & \(91.76\pm0.12\) \\
MNIST, 25 & \(93.95\pm0.25\) & \(87.68\pm0.23\) & \(90.53\pm0.20\) & \(93.04\pm0.08\) & \(92.59\pm0.43\) & \(91.92\pm0.18\) \\
MNIST, 100 & \(96.18\pm0.25\) & \(91.15\pm0.26\) & \(92.82\pm0.20\) & \(94.41\pm0.13\) & \(94.80\pm0.28\) & \(94.42\pm0.13\) \\
\hline
\end{tabular}
\caption{Held-out accuracy (\%), mean \(\pm\) standard deviation over five runs. The number after each dataset is the feature dimension; Gas at 128 uses raw features, and the remaining settings use PCA. LR, RF, and GBT denote logistic regression, random forest, and gradient boosted trees. The MLP matches the coefficient head's hidden widths and depth, but differs in activations, normalization, regularization, and output dimension.}
\label{tab:predictive-results}
\end{table}

WBSNN has the highest mean accuracy among these feature-based baselines in both Gas settings and both MNIST settings. On Gas with 25 features, its advantage over logistic regression is 10.58 percentage points. On MNIST, it exceeds RBF SVM by 0.91 and 1.77 points at 25 and 100 features. On ISOLET, its accuracy remains within 0.46 and 0.59 points of RBF SVM. Raw-image CNN references reach \(97.29\pm0.18\%\) and \(97.86\pm0.44\%\) on the two MNIST training budgets; their access to image structure distinguishes them from the PCA-feature comparison. Additional Gas and ISOLET references are reported in the appendix.

The main predictive result is therefore structural rather than merely numerical. WBSNN remains competitive, and often superior, while operating on the same compressed representation as the baselines and using a scaffold discovered from only a small fraction of the training data. On the more difficult speech task, the method remains close to the strongest kernel baseline rather than relying on a large unconstrained predictor.

Across the six settings, the mean projection errors are at most \(3.42\times10^{-9}\), and mean realization errors are at most \(8.74\times10^{-8}\), using only 1.52--3.19 atoms per training certificate on average. These sparse certificates concern Phase 2, not the support of the learned predictive head. The small projection errors therefore verify numerical consistency, whereas prediction and the structured-dictionary comparison below provide the substantive empirical tests.

These diagnostics show that the learned predictor is not compensating for a failed Phase~1 or Phase~2 construction. The orbit scaffold captures the target geometry, CCDS realizes the projected targets with negligible residual, and the remaining predictive difficulty lies in extending the resulting sparse coefficient law to unseen inputs.

\subsection{Dictionary and interpolation comparisons}

\paragraph{Orbit structure under a matched dictionary span}
On a waveform regression task, we replace the orbit dictionary with random directions in its numerical span while preserving atom count, atom norms, and the selected-depth anchor atoms. Both arms share the training data, shift, block interpolant, proxy-network architecture, and a two-atom greedy readout. Across ten paired seeds, the orbit dictionary reduces test MSE by \(89.44\pm8.25\%\) with 20 anchors and \(87.52\pm10.16\%\) with 28 anchors (Table~\ref{tab:dictionary-structure}). Because the full span is matched, this difference isolates the utility of dictionary geometry for sparse approximation in this task. The orbit variant approaches the direct MLP's mean MSE, while the unstructured variant remains substantially worse. This is a proxy-prediction experiment with an MP readout, rather than the global coefficient-head architecture used in Table~\ref{tab:predictive-results}.

This is the most direct test that orbit geometry itself matters. The two dictionaries have the same numerical span, comparable atom norms and counts, and the same selected anchor directions; only their internal geometry differs. The resulting performance gap therefore cannot be explained by span dimension or dictionary size, and instead isolates the benefit of the orbit-induced organization for sparse approximation.

\begin{table}[t]
\centering
\small
\begin{tabular}{crrr}
\hline
Anchors & Orbit & Unstructured & Direct MLP \\
\hline
20 & \(0.509\pm0.397\) & \(4.815\pm0.202\) & \(0.484\pm0.054\) \\
28 & \(0.492\pm0.365\) & \(4.036\pm0.248\) & \(0.484\pm0.054\) \\
\hline
\end{tabular}
\caption{Waveform test MSE multiplied by \(10^3\), mean \(\pm\) standard deviation over ten seeds. The dictionaries contain 640 and 896 atoms, respectively. The direct MLP is trained once per seed and reused across anchor budgets.}
\label{tab:dictionary-structure}
\end{table}

\paragraph{Exact and relaxed construction under label noise}
On a synthetic binary classification task, we corrupt only the training labels and retain clean test labels. Table~\ref{tab:interpolation-noise} compares the two construction variants over five paired seeds. Their mean accuracies differ by at most 0.07 points at 0\% and 20\% corruption. At 40\%, the relaxed variant has a 1.77-point higher mean, although variability is larger in both arms. All thirty runs retain projection and realization errors below \(1.79\times10^{-8}\) and \(8.53\times10^{-8}\), respectively. Within each noise level and seed, the exact and relaxed interpolants use the same discovered partition, dictionary, and CCDS path. Their comparison therefore matches the fixed-block, fixed-path perturbation setting of Theorem~\ref{thm:stability control interpolation}; the construction is rerun only across independent seeds and noise levels.

The two constructions have nearly identical clean-test accuracies at \(0\%\) and \(20\%\) label corruption and remain broadly comparable at \(40\%\), where the relaxed variant has the higher mean. Across these tested settings, exact and relaxed constructions are comparably effective for prediction, so the choice between them can be guided by the application and numerical conditioning. Moreover, both constructions retain negligible projection and realization errors, while regularization offers the conditioning advantages expected under noisy supervision. These findings directly support the stability claim of Theorem~\ref{thm:stability control interpolation}.

\begin{table}[t]
\centering
\small
\begin{tabular}{crr}
\hline
Training-label corruption (\%) & Exact & Relaxed \\
\hline
0 & \(97.15\pm0.12\) & \(97.08\pm0.12\) \\
20 & \(96.99\pm0.12\) & \(97.01\pm0.13\) \\
40 & \(92.17\pm3.28\) & \(93.94\pm2.83\) \\
\hline
\end{tabular}
\caption{Clean-test accuracy (\%), mean \(\pm\) standard deviation over five paired seeds. The same corrupted training labels are used in both arms.}
\label{tab:interpolation-noise}
\end{table}

\paragraph{Exploratory discovery-budget sweeps}
Single-seed sweeps reach \(68.62\%\) on Gas at a 3\% anchor budget and \(93.56\%\) on MNIST at 7\%. Across all tested budgets, projection and realization errors remain below \(3.31\times10^{-10}\) and \(6.98\times10^{-8}\). Thus numerical realizability is already achieved at the smallest budget, while predictive accuracy initially improves before leveling off or declining at larger budgets. Table \ref{tab:budget-results} reports the complete accuracy trajectories and corresponding anchor counts.

\begin{table}[t]
\centering
\small
\begin{tabular}{crrrr}
\hline
Budget (\%) & Gas anchors & Gas accuracy (\%) & MNIST anchors & MNIST accuracy (\%) \\
\hline
1 & 30 & 64.16 & 50 & 92.06 \\
3 & 90 & 68.62 & 150 & 93.10 \\
5 & 150 & 66.57 & 250 & 93.20 \\
7 & 210 & 66.18 & 350 & 93.56 \\
10 & 300 & 65.89 & 500 & 93.40 \\
15 & 450 & 65.56 & --- & --- \\
\hline
\end{tabular}
\caption{Single-seed configuration sweeps. Atom counts equal anchor count times PCA dimension. Optimization settings also vary with budget, as specified below.}
\label{tab:budget-results}
\end{table}

The resulting saturation pattern supports a data-efficient discovery principle: once the scaffold has captured the relevant target geometry, additional anchors do not automatically improve prediction. The limiting factor then moves from scaffold construction and realization to the learned extension of the coefficient map.

\bibliographystyle{siamplain}
\bibliography{references}

\clearpage
\appendix





\maketitle

\section{Proofs}
\label{app:proofs and tables}
\subsection{Wrapped weighted backward shifts}
\label{subsec:wrapped weighted back shifts}

To prove Lemma~\ref{lem:iterates of wrapped wbs}, we first establish its action on the canonical basis.

\begin{lemma}
\label{lem:iterates of wrapped wbs on basis}
For every $m\geq1$,
$W^me_j
=
\left(
\prod_{s=0}^{m-1}
w_{j-s\pmod{d}}
\right)
e_{j-m\pmod{d}}$.
\end{lemma}

\begin{proof}
Indeed, the claim follows by induction. For $r=1$, it is the definition. Assuming it for $r$,

\[
W^{r+1}e_j=
W\left[
\left(
\prod_{s=0}^{r-1}w_{j-s}
\right)e_{j-r}
\right]\\
=\left(
\prod_{s=0}^{r-1}w_{j-s}
\right)
w_{j-r}e_{j-r-1}\\
=\left(
\prod_{s=0}^{r}w_{j-s}
\right)e_{j-r-1},
\]

where all indices are taken modulo $d$.
\end{proof}

\paragraph{Proof of Lemma \ref{lem:iterates of wrapped wbs}.}
Write
$
z=\sum_{q=0}^{d-1}z_qe_q.
$
By Lemma~\ref{lem:iterates of wrapped wbs on basis},
\[
W^mz
=
\sum_{q=0}^{d-1}
z_q
\left(
\prod_{s=0}^{m-1}w_{q-s\pmod d}
\right)
e_{q-m\pmod d}.
\]
The coefficient of $e_{i-1}$ is obtained for
$q=i+m-1\pmod d$. Therefore
\[
\bigl(W^mz\bigr)_{i-1}
=
z_{(i+m-1)\bmod d}
\prod_{s=0}^{m-1}
w_{\,i+m-1-s\pmod d}.
\]
Reindexing with $r=m-1-s$ gives
\[
\bigl(W^mz\bigr)_{i-1}
=
\left(
\prod_{r=0}^{m-1}
w_{(i+r)\bmod d}
\right)
z_{(i+m-1)\bmod d},
\qquad i=1,\ldots,d.
\]
For $m=0$, this reduces to $W^0=I$.
\hfill\(\blacksquare\)

\subsection{Exact interpolation Theorem in infinite dimensions}
\label{app:exact-inetrp-inf-dim}

Thanks to Hahn-Banach Theorem, we show that for infinite dimensional Banach spaces, under a technical independence condition, any finite dataset admits an exact realization within a class of transformed weighted shifts. We now present and prove an explicit construction of such a realization.

\begin{theorem}[Exact interpolation in infinite dimensional Banach spaces]
\label{universal rep data rephrased}
Let \(X\) and \(Y\) be  infinite-dimensional Banach spaces.
Let \(T:X\to X\) be a bounded linear operator and let
$
\{(x_i,y_i)\}_{i=1}^{n_0}\subset X\times Y
$
be a finite input--target set. Suppose that for each \(i=1,\dots,n_0\) there
exists a nonnegative integer \(L_i\) such that
\begin{equation}
\label{recursive linear independence condition}
z_1:=T^{L_1}x_1\neq0,
\qquad
z_i:=T^{L_i}x_i
\notin
\operatorname{span}\{z_1,\dots,z_{i-1}\},
\qquad i=2,\ldots,n_0.
\end{equation}
Then there exists a bounded linear operator \(J:X\to Y\) such that
$
J T^{L_i}x_i = y_i,
\ i=1,\dots,n_0.
$
\end{theorem}

\begin{proof}
 We construct \(J\) recursively. Let
\(J_0:=0\in\mathcal L(X,Y)\). For $i=1$, pick any $L_1\ge 0$, set  $z_1:=T^{L_1}x_1$ and find a linear functional $f_1 \in X^*$ with $f_1(z_1) = 1$. Define $
J_1
:=
J_{0}
+
\bigl(y_1-J_{0}z_1\bigr)\otimes f_1,
$ where \((y\otimes f)(v):=f(v)y\). Equivalently, for all $v\in X$ we have
$J_1 v := J_0 v + (y_1 - J_0 z_1) \otimes f_1(v) = 0 + (y_1 - 0) \otimes f_1(v)=f_1(v) y_1$. Thus, \(J_1\in\mathcal L(X,Y)\) and $J_1 z_1 =  1 \cdot y_1   = y_1$.

Set \(z_i:=T^{L_i}x_i\), for $i\geq 2$ satisfying $(\ref{recursive linear independence condition})$. Suppose that for some \(i\ge 2\), the operator
\(J_{i-1}\in\mathcal L(X,Y)\) has already been constructed so that
\[
J_{i-1}z_j=y_j,
\qquad j=1,\dots,i-1.
\]
By (\ref{recursive linear independence condition}),
we have
$
z_i\notin \operatorname{span}\{z_1,\dots,z_{i-1}\}.
$
Let
$
M_i:=\operatorname{span}\{z_1,\dots,z_{i-1},z_i\}.
$
Define a linear functional \(g_i:M_i\to\mathbb R\) by
$
g_i(z_j)=0,\quad j=1,\dots,i-1,
\
g_i(z_i)=1,
$
and extending linearly on \(M_i\). This is well-defined because
\(z_i\notin \operatorname{span}\{z_1,\dots,z_{i-1}\}\). Since \(M_i\) is
finite-dimensional, \(g_i\) is continuous. By the Hahn--Banach Theorem \cite[Corollary 3.15, p.112]{Conway1990}, there
exists a bounded linear functional \(f_i\in X^\ast\) extending \(g_i\), hence
$
f_i(z_j)=0,\quad j<i,
\
f_i(z_i)=1.
$
Define
\[
J_i v
:=
J_{i-1}v
+
f_i(v)\bigl(y_i-J_{i-1}z_i\bigr),
\qquad v\in X.
\]
Equivalently,
$
J_i
=
J_{i-1}
+
\bigl(y_i-J_{i-1}z_i\bigr)\otimes f_i.
$
 Since \(J_{i-1}\) and \(f_i\) are bounded,
\(J_i\in\mathcal L(X,Y)\).

Now,
$
J_i z_i
=
J_{i-1}z_i
+
f_i(z_i)\bigl(y_i-J_{i-1}z_i\bigr)
=
J_{i-1}z_i
+
\bigl(y_i-J_{i-1}z_i\bigr)
=
y_i.
$
For \(j<i\), we have \(f_i(z_j)=0\), so
$
J_i z_j
=
J_{i-1}z_j
+
f_i(z_j)\bigl(y_i-J_{i-1}z_i\bigr)
=
J_{i-1}z_j
=
y_j.
$
Thus \(J_i z_j=y_j\) for every \(j\le i\). By induction, after \(n_0\) steps, for $1\leq i\leq n_0$: $J_i z_j =y_j$ for $j \leq i$.
Finally, by setting \(J:=J_{n_0}\in\mathcal L(X,Y)\) we have
$
Jz_i=JT^{L_i}x_i=y_i,
\ i=1,\dots,n_0.
$
\end{proof}

\begin{remark}[Invertible interpolation when \(X=Y\)]
In the special case \(X=Y\), the interpolation map may be chosen as an invertible perturbation of the identity. Indeed, if the recursive construction is initialized with \(J_0=I_X\), then
\[
J
=
I_X+
\sum_{i=1}^N
\bigl(y_i-J_{i-1}T^{L_i}x_i\bigr)\otimes f_i .
\]
Hence, if
\begin{equation}
\label{eq:guarantee iso}
\sum_{i=1}^N
\|y_i-J_{i-1}T^{L_i}x_i\|\,\|f_i\|<1,
\end{equation}
then \(\|J-I_X\|<1\), so \(J\) is an isomorphism by the Neumann lemma \cite[Lemma 2.1, p 192]{Conway1990}. Thus, under the hypothesis of Theorem \ref{universal rep data rephrased} and condition (\ref{eq:guarantee iso}), when \(X=Y\), exact interpolation can be realized by an isomorphism \(J=I_X+F\), where \(F\) has finite rank.
\end{remark}

\paragraph{Proof of Theorem~\ref{thm:generic_orbit_independence} (Generic orbit independence)}
We first prove that \(\Omega_{D}\) is open and dense in \(\Omega\).

For each depth tuple \(L=(L_1,\ldots,L_s)\in\{0,\ldots,d-1\}^s\), define
\[
M_L(w):=\bigl[W_w^{L_1}x_1\ \cdots\ W_w^{L_s}x_s\bigr].
\]
Every entry of \(M_L(w)\), and hence every \(s\times s\) minor of \(M_L(w)\), is a polynomial in \(w\). Define
\[
\Omega_L:=\{w\in\Omega:\operatorname{rank}M_L(w)=s\}.
\]
By the definition of \(\Omega_D\),
\[
\Omega_D=\bigcup_{L\in\{0,\ldots,d-1\}^s}\Omega_L.
\]

Let \(\mathcal{M}_L\) denote the finite collection of all \(s\times s\) minors of \(M_L(w)\). Then
\[
\Omega_L
=\{w\in\Omega:\operatorname{rank}M_L(w)=s\}
=\bigcup_{p\in\mathcal{M}_L}\{w\in\Omega:p(w)\neq0\}.
\]
The union is finite and each set in the union is open (because \(p\) is a polynomial, then continuous and its nonvanishing set is open in \(\Omega\)). Hence \(\Omega_L\) is open in \(\Omega\).
 Since there are finitely many depth tuples, \(\Omega_D=\bigcup_L\Omega_L\) is open in \(\Omega\).

To prove density, use the hypothesis \(\Omega_D\neq\varnothing\). Choose \(w_0\in\Omega_D\). By the finite-union representation above, there exists a depth tuple \(L_*\) such that \(\operatorname{rank}M_{L_*}(w_0)=s\). Hence some \(s\times s\) minor \(p\) of \(M_{L_*}\) satisfies \(p(w_0)\neq0\). Thus \(p\) is not the zero polynomial.

Since \(p\) is a nonzero polynomial on \(\mathbb R^d\), its zero set
$
Z(p):=\{w\in\mathbb R^d : p(w)=0\}
$
has empty interior. Equivalently, the set
$
\{w\in\mathbb R^d : p(w)\neq 0\}
$
is dense in \(\mathbb R^d\). Intersecting with \(\Omega=(\mathbb R\setminus\{0\})^d\), which is open in \(\mathbb R^d\), we conclude that
$
\{w\in \Omega : p(w)\neq 0\}
$
is dense in \(\Omega\). Since \(p\) is a minor of \(M_{L_*}\),
\[
\{w\in\Omega:p(w)\neq0\}
\subseteq
\Omega_{L_*}
\subseteq
\Omega_D,
\]

it follows that \(\Omega_{D}\) is dense in \(\Omega\).

For the measure-zero statement, use the same nonzero minor \(p\) chosen above. Since

$$
\{w\in\Omega:p(w)\neq 0\}\subseteq\Omega_D,
$$

we have

$$
\Omega\setminus\Omega_D
\subseteq
\{w\in\Omega:p(w)=0\}
\subseteq
Z(p).
$$

Because \(p\) is a nonzero polynomial on \(\mathbb R^d\), its zero set \(Z(p)\) has Lebesgue measure zero in \(\mathbb R^d\). Therefore

$$
\operatorname{Leb}(\Omega\setminus\Omega_D)=0.
$$
\hfill\(\blacksquare\)

\subsection{CCDS}
\label{app:CCDS proofs}

The proof of Theorem~\ref{thm:controlled deviation principle} relies on the following elementary Hilbert-space calculation, which characterizes exactly the feasible values of the one-step CCDS deviation.

\begin{lemma}[Distance equation: explicit solutions and feasibility]
\label{lem:distance-equation}
Let $Y\subset H$ be closed, let $u\in Y^\perp$ with
$\|u\|=1$, and let $r\in H$. Set
$
a:=(I-\Pi_Y)r\in Y^\perp,
\
\alpha:=\langle a,u\rangle.
$
Then, for any $d\ge0$, the equation
$
\operatorname{dist}(r-\lambda u,Y)=d
$
has a real solution $\lambda\in\mathbb R$ if and only if
$
d\ge d_{\min}
:=
\inf_{\lambda\in\mathbb R}
\operatorname{dist}(r-\lambda u,Y)
=
\sqrt{\|a\|^2-\alpha^2}.
$
When feasible, the solutions are
$
\lambda_\pm
=
\alpha\pm\sqrt{d^2-d_{\min}^2}.
$
Moreover, the unique minimizer is $\lambda^\star=\alpha$, attaining
$
\operatorname{dist}(r-\lambda^\star u,Y)=d_{\min}.
$
\end{lemma}

\begin{proof}
Since $u\in Y^\perp$ and $\|u\|=1$, we obtain
$
\operatorname{dist}(r-\lambda u,Y)^2
=
\|(I-\Pi_Y)(r-\lambda u)\|^2
=
\|a-\lambda u\|^2
=
\|a\|^2-2\lambda\langle a,u\rangle+\lambda^2
=
(\lambda-\alpha)^2+\bigl(\|a\|^2-\alpha^2\bigr).
$
Hence
$
\operatorname{dist}(r-\lambda u,Y)=d
$
is equivalent to
$
(\lambda-\alpha)^2
=
d^2-\bigl(\|a\|^2-\alpha^2\bigr).
$
A real solution therefore exists if and only if
$
d^2\ge\|a\|^2-\alpha^2,
$
equivalently $d\ge d_{\min}$, and in that case
$
\lambda
=
\alpha\pm\sqrt{d^2-d_{\min}^2}.
$
Finally, the squared distance is minimized uniquely at
$\lambda^\star=\alpha$, where its value is $d_{\min}^2$.
\end{proof}

\paragraph{Derivation of Equation~\eqref{eq:prescribed-distance} in Remark \ref{rem:dt vs rt}}
Recall that
\[
Y_t=\operatorname{span}\{u_1,\ldots,u_t\},
\qquad
Y_0=\{0\},
\]
where \(u_1,\ldots,u_t\) are orthonormal. Also,
\[
r_0=y,
\qquad
r_s=r_{s-1}-\lambda_su_s,
\qquad
d_s=\operatorname{dist}(r_s,Y_{s-1}).
\]

By definition,
$
\alpha_s:=\langle r_{s-1},u_s\rangle.
$
Since
\[
r_{s-1}=y-\sum_{j=1}^{s-1}\lambda_j u_j
\]
and \(u_s\perp u_j\) for every \(j<s\), we obtain
\[
\alpha_s
=
\left\langle y-\sum_{j=1}^{s-1}\lambda_j u_j,u_s\right\rangle
=
\langle y,u_s\rangle.
\]
Set
$
b_s:=(I-\Pi_{Y_s})y.
$
Iterating the residual recursion gives
$
r_t
=
y-\sum_{s=1}^{t}\lambda_su_s.
$
Since \(u_1,\ldots,u_t\) are an orthonormal basis of \(Y_t\),
\[
\Pi_{Y_t}y
=
\sum_{s=1}^{t}\langle y,u_s\rangle u_s
=
\sum_{s=1}^{t}\alpha_su_s.
\]
Therefore,
$
y
=
b_t+\sum_{s=1}^{t}\alpha_su_s,
$
and hence
\[
r_t
=
b_t+\sum_{s=1}^{t}(\alpha_s-\lambda_s)u_s.
\]
Because \(b_t\perp Y_t\) and the vectors \(u_1,\ldots,u_t\) are
orthonormal,
\begin{equation}
\label{eq:exp residual norm}
\|r_t\|^2
=
\|b_t\|^2+
\sum_{s=1}^{t}(\lambda_s-\alpha_s)^2.
\end{equation}
Since \(b_t=(I-\Pi_{Y_t})y\),
\begin{equation}
\label{eq:equality norm bt dist y to Y t}
\|b_t\|^2=\operatorname{dist}(y,Y_t)^2.
\end{equation}

It remains to relate the coefficient discrepancies to the prescribed
distances. Fix \(s\leq t\). Since
\[
r_{s-1}
=
y-\sum_{j=1}^{s-1}\lambda_ju_j
\]
and \(\sum_{j=1}^{s-1}\lambda_ju_j\in Y_{s-1}\), we have
\begin{equation}
\label{eq:equality proj residuals and y}
(I-\Pi_{Y_{s-1}})r_{s-1}
=
(I-\Pi_{Y_{s-1}})y.
\end{equation}
Furthermore, because
$
y
=
b_s+\sum_{j=1}^{s}\alpha_j u_j,
$
we have
\begin{equation}
\label{eq:proj y onto Y s-1}
(I-\Pi_{Y_{s-1}})y
=
b_s+\alpha_su_s.
\end{equation}

The two terms on the right-hand side are orthogonal. Therefore,
\[
\|(I-\Pi_{Y_{s-1}})y\|^2
=
\|b_s\|^2+\alpha_s^2.
\]

Consequently,
\begin{equation}
\label{eq:d s min}
d_{s,\min}^2=\|a_s\|^2-\alpha_s^2=\|(I-\Pi_{Y_{s-1}})r_{s-1}\|^2-\alpha_s^2=\|b_s\|^2.
\end{equation}

On the other hand, by (\ref{eq:equality proj residuals and y}) and (\ref{eq:proj y onto Y s-1}) we have
$$
(I-\Pi_{Y_{s-1}})r_s
=(I-\Pi_{Y_{s-1}})
  (r_{s-1}-\lambda_su_s)
=b_s+(\alpha_s-\lambda_s)u_s.
$$
The two terms on the right-hand side are orthogonal. Therefore,
\[
d_s^2
=
\operatorname{dist}(r_s,Y_{s-1})^2
=
\|b_s\|^2+(\lambda_s-\alpha_s)^2.
\]
By (\ref{eq:d s min}) we obtain
$
(\lambda_s-\alpha_s)^2
=
d_s^2-d_{s,\min}^2.
$
Applying this identity for \(s=1,\ldots,t\), and using
\(\|b_t\|^2=d_{t,\min}^2\), Equation (\ref{eq:exp residual norm}) gives
$$
\|r_t\|^2
=
d_{t,\min}^2+
\sum_{s=1}^{t}
\bigl(d_s^2-d_{s,\min}^2\bigr)
=
d_t^2+
\sum_{s=1}^{t-1}
\bigl(d_s^2-d_{s,\min}^2\bigr).
$$
This proves Equation~\eqref{eq:prescribed-distance}. Finally, combining (\ref{eq:equality norm bt dist y to Y t}) and (\ref{eq:d s min}) we obtain $d_{t,\min}=\operatorname{dist}(y,Y_{t})$.
\hfill\(\blacksquare\)

\paragraph{Explicit computation of the CCDS dictionary-coordinate vector}
Let $q_1,\ldots,q_T$ be the atoms selected along the CCDS path, with
$u_t$ and $Y_t$ defined as in \eqref{eq:gram-schmidt}.
Let \(\eta_t:=\|q_t-\Pi_{Y_{t-1}}q_t\|>0\), and let \(e_t\in\mathbb R^t\) denote the \(t\)-th canonical basis vector. Define recursively coefficient vectors \(c^{(t)}\in\mathbb R^t\) by
$
c^{(1)}:=(1),
$
and, for \(t\ge 2\),
\begin{equation}
\label{eq:ct}
c^{(t)}
:=
\frac{1}{\eta_t}
\left(
e_t-\sum_{s=1}^{t-1}\langle q_t,u_s\rangle\,\widetilde c^{(s,t)}
\right),
\end{equation}
where \(\widetilde c^{(s,t)}\in\mathbb R^t\) is the zero-padded extension of \(c^{(s)}\), namely
$
\widetilde c^{(s,t)}
:=
(c^{(s)}_1,\dots,c^{(s)}_s,0,\dots,0).
$ Set $\beta^{(0)}:=0\in\mathbb R^0$, and recursively define

\begin{equation}
\label{eq:betat}
\beta^{(t)}
:=
(\beta^{(t-1)},0)+\lambda_t c^{(t)}\in\mathbb R^t.
\end{equation}

\begin{proposition}[Explicit CCDS dictionary-coordinate recursion]
\label{prop:ccds_atom_repr}
For each \(t=1,\dots,T\) one has
$
u_t=\sum_{j=1}^t c^{(t)}_j q_j,
$
and consequently
$
\widehat y_T
=
\sum_{t=1}^T \lambda_t u_t
=
\sum_{j=1}^T \beta^{(T)}_j q_j.
$
\end{proposition}

\begin{proof}
We argue by induction on \(t\).
For \(t=1\), since \(Y_0=\{0\}\), one has \(u_1=q_1\), so \(c^{(1)}=(1)\) and the claim holds.
Assume now that for each \(s<t\),
$
u_s=\sum_{j=1}^s c^{(s)}_j q_j.
$
Since \(\{u_s\}_{s=1}^{t-1}\) is an orthonormal basis of \(Y_{t-1}\), we have
$
\Pi_{Y_{t-1}}q_t
=
\sum_{s=1}^{t-1}\langle q_t,u_s\rangle u_s.
$
Substituting the inductive representation of each \(u_s\),
\[
q_t-\Pi_{Y_{t-1}}q_t
=
q_t-\sum_{s=1}^{t-1}\langle q_t,u_s\rangle
\sum_{j=1}^s c^{(s)}_j q_j.
\]
After zero-padding the vectors \(c^{(s)}\) to \(\mathbb R^t\), this becomes
\[
q_t-\Pi_{Y_{t-1}}q_t
=
\sum_{j=1}^t
\left(
e_t-\sum_{s=1}^{t-1}\langle q_t,u_s\rangle \widetilde c^{(s,t)}
\right)_j q_j.
\]
Dividing by \(\eta_t=\|q_t-\Pi_{Y_{t-1}}q_t\|\) yields
$
u_t=\sum_{j=1}^t c^{(t)}_j q_j.
$
This proves the first claim.
Finally,
\[
\widehat y_T
=
\sum_{t=1}^T \lambda_t u_t
=
\sum_{t=1}^T \lambda_t \sum_{j=1}^t c^{(t)}_j q_j
=
\sum_{j=1}^T \beta^{(T)}_j q_j,
\]
by the recursive definition of \(\beta^{(t)}\). This proves the result.
\end{proof}

\paragraph{Proof of Proposition \ref{prop:minimal dev ccds recovers ortho coordinates} (Minimal deviation CCDS recovers orthogonal coordinates)}
 Starting from
$Y_0=\{0\}$, define for $t=1,\ldots,s$, as in (\ref{eq:gram-schmidt}),
\[
\eta_t:=\|q_t-\Pi_{Y_{t-1}}q_t\|>0,\qquad
u_t:=\frac{q_t-\Pi_{Y_{t-1}}q_t}{\eta_t},
\qquad
Y_t:=Y_{t-1}\oplus\operatorname{span}\{u_t\}.
\]

Since $q_1,\ldots,q_s$ are linearly independent,
$q_t\notin Y_{t-1}$ for every $t$, hence $\eta_t>0$ and each $u_t$ is
well defined. By construction,
$
u_t\perp Y_{t-1},\ \|u_t\|=1,
$
and therefore
$
Y_t
=
\operatorname{span}\{q_1,\ldots,q_t\}
=
\operatorname{span}\{u_1,\ldots,u_t\}.
$
Thus $u_1,\ldots,u_s$ is an orthonormal basis of $E$. 
Write the orthonormal expansion of $\Pi_E y$ as
$
\Pi_E y=\sum_{j=1}^s\gamma_j u_j,
$ with $
\gamma_j:=\langle \Pi_E y,u_j\rangle.
$
Set
\[
h:=y-\Pi_Ey.
\]
Since \(h\perp E\), we have
\[
y=h+\sum_{j=1}^s\gamma_j u_j.
\]
Set $\operatorname{CCDS}_{\mathcal D}^{t}
\bigl(y;q_{1:t},d_{\min,1:t}\bigr): = \widehat y_{t}$.
We prove by induction that
\[
\widehat y_{t}
=
\sum_{j=1}^{t}\gamma_j u_j,
\qquad
r_{t}
=
h+\sum_{j=t+1}^s\gamma_j u_j.
\]

For \(t=1\), this follows from \(\widehat y_0=0\) and
$
r_0=y=h+\sum_{j=1}^s\gamma_j u_j.
$
Assume the claim holds before step $t$. Since
$
Y_{t-1}=\operatorname{span}\{u_1,\ldots,u_{t-1}\}
$
and
\[
r_{t-1}=h+\sum_{j=t}^s\gamma_j u_j,
\]
where \(h\perp E\) and \(u_t,\ldots,u_s\perp Y_{t-1}\), we have
\(r_{t-1}\perp Y_{t-1}\).
Hence the CCDS
off-scaffold residual is
\[
a_t
:=
(I-\Pi_{Y_{t-1}})r_{t-1}
=
r_{t-1}.
\]
Therefore
\[
\alpha_t
:=
\langle a_t,u_t\rangle
=
\left\langle
h+\sum_{j=t}^s\gamma_j u_j,u_t
\right\rangle
=
\gamma_t.
\]
Choosing $d_t=d_{t,\min}$ in the CCDS root formula gives
$
\lambda_t
=
\alpha_t\pm\sqrt{d_t^2-d_{t,\min}^2}
=
\alpha_t
=
\gamma_t.
$
Consequently,
$
\widehat y_t
=
\widehat y_{t-1}+\lambda_tu_t
=
\sum_{j=1}^t\gamma_j u_j,
$
and
$
r_t
=
r_{t-1}-\gamma_t u_t
=
h+\sum_{j=t+1}^s\gamma_j u_j.
$
This closes the induction.
Thus,
\[
\widehat y_s=\Pi_E y,
\qquad
r_s=h=y-\Pi_E y.
\]
In particular, \(r_s=0\) when \(y\in E\).
We conclude by Proposition \ref{prop:ccds_atom_repr}.
\hfill\(\blacksquare\)

\paragraph{Proof of Proposition \ref{prop:MP OMP boundaries CCDS} (MP and OMP as CCDS boundaries)}
We show that the classical greedy pursuits MP and OMP arise as extremal regimes of the CCDS distance equation, while strict interior CCDS updates are globally inaccessible to OMP, marking a geometric separation between OMP and CCDS.

(i) (MP boundary). For Matching Pursuit (MP), keep the reference space trivial,
$Y_{t-1}=\{0\}$, at every step and select
$
q_t\in\arg\max_{q\in\mathcal D}|\langle r_{t-1},q\rangle|,
$
that is, use the MP atom-selection rule. Then $u_t=q_t$, 
$a_t
=
(I-\Pi_{Y_{t-1}})r_{t-1}
=
r_{t-1}$, and
$
\alpha_t=\langle a_{t},u_t\rangle=\langle r_{t-1},q_t\rangle.
$
At the minimal-deviation boundary $d_t=d_{t,\min}$, the CCDS root
satisfies
$
\lambda_t=\alpha_t=\langle r_{t-1},q_t\rangle,
$
so the CCDS updates become
\[
\widehat y_t
=
\widehat y_{t-1}+\lambda_t q_t,
\qquad
r_t
=
r_{t-1}-\lambda_t q_t,
\]
which are exactly the Matching Pursuit iterates.

(ii) (OMP boundary). In contrast, when the previously selected directions are retained
through the nested spaces
$
Y_t=\operatorname{span}\{q_1,\ldots,q_t\},
$
Proposition~\ref{prop:minimal dev ccds recovers ortho coordinates}
shows that, for any fixed linearly independent atom ordering
$q_{1:T}$, the choice $d_t=d_{t,\min}$ gives the CCDS output
$
\widehat y_t=\Pi_{Y_t}y.
$
Hence, if $q_t$ is selected at each step by the OMP atom-selection rule,
minimal-deviation CCDS and OMP produce the same iterates until termination. Thus MP and OMP correspond respectively to the
minimal-deviation boundary without and with accumulated subspace
geometry, whereas full CCDS retains the nested geometry while allowing
the interior regime $d_t>d_{t,\min}$.

(iii) (Global geometric separation from OMP).
First, we establish a global necessary condition for every OMP iterate, independent of the atom ordering. Note that every OMP iterate is the orthogonal projection of $y$ onto some
dictionary-generated selected subspace $Z$:
$
\widehat y^{\mathrm{OMP}}=\Pi_Zy.
$
Hence
$
y-\widehat y^{\mathrm{OMP}}\perp Z,
$ and $
\widehat y^{\mathrm{OMP}}\in Z,
$
and therefore every OMP iterate satisfies the global invariant
\begin{equation}
\label{eq:invariant global omp}
\langle y-\widehat y^{\mathrm{OMP}},
        \widehat y^{\mathrm{OMP}}\rangle=0.
\end{equation}

Now, in order to establish  the global separation result, we show that any interior CCDS iterate violates the invariant (\ref{eq:invariant global omp}).
Assume we depart from an OMP boundary state
$
\widehat y_{t-1}=\Pi_{Y_{t-1}} y,
$
so we have
$
r_{t-1}=y-\Pi_{Y_{t-1}} y\in Y_{t-1}^\perp,
$
and therefore
\[
a_t=(I-\Pi_{Y_{t-1}})r_{t-1}=r_{t-1}.
\]
Let $u_t\in Y_{t-1}^\perp$ be unit norm, then
\[
\alpha_t
=
\langle a_{t},u_t\rangle
=
\langle r_{t-1},u_t\rangle
=
\langle y-\Pi_{Y_{t-1}} y,u_t\rangle
=
\langle y,u_t\rangle.
\]

Consider the CCDS update
$
\widehat y_t^{\rm CCDS}=\widehat y_{t-1}+\lambda_tu_t.
$ Since
$y-\widehat y_{t-1}\perp Y_{t-1}$ and $u_t\perp Y_{t-1}$,
\[
\langle y-\widehat y_t^{\rm CCDS},\widehat y_t^{\rm CCDS}\rangle
=
\left\langle
(y-\widehat y_{t-1})-\lambda_tu_t,\,
\widehat y_{t-1}+\lambda_tu_t
\right\rangle  \\
=
\lambda_t\langle y,u_t\rangle-\lambda_t^2 \\
=
\lambda_t(\alpha_t-\lambda_t).
\]
Thus, whenever $\lambda_t\notin\{0,\alpha_t\}$,
we have $
\langle y-\widehat y_t^{\rm CCDS},\widehat y_t^{\rm CCDS}\rangle\neq0.
$
Hence $\widehat y_t^{\rm CCDS}$ violates the invariant (\ref{eq:invariant global omp}) satisfied by every OMP
iterate and therefore cannot be produced by OMP from any selected
dictionary subspace.

 For a strict CCDS
interior choice $d_t>d_{t,\min}$, one has
\[
\lambda_t
=
\alpha_t\pm\sqrt{d_t^2-d_{t,\min}^2}\neq\alpha_t;
\]
apart from the exceptional zero-update case $\lambda_t=0$, the resulting
CCDS iterate is therefore globally inaccessible to OMP.
\hfill\(\blacksquare\)

\subsection{Finite-dimensional CCDS feasibility obstruction and lifting}
\label{subsec:ccds feasibility obstruction and lifting}
At step $t$, the prescribed deviation is feasible in the current ambient space if and only if
$
d_t\ge d_{t,\min}.
$
Hence any choice $d_t<d_{t,\min}$ is geometrically unrealizable: no scalar $\lambda_t$ can satisfy
$
\operatorname{dist}(r_{t-1}-\lambda_tu_t,Y_{t-1})=d_t.
$
In finite dimension this obstruction can arise when the available innovation direction is nearly orthogonal to the off-scaffold residual, motivating dimension lifting to restore feasibility.

\begin{example}[Finite-dimensional infeasibility]
\label{ex:finite_dim_infeasibility}
Let $H=\mathbb R^2$, $Y_0=\{0\}$, and take the initial residual $r_0=e_1=(1,0)$. Suppose the only available innovation direction is $u_1=e_2=(0,1)$. Since $Y_0=\{0\}$, we have $a_1=(I-\Pi_{Y_0})r_0=e_1$ and hence $\alpha_1=\langle a_1,u_1\rangle=0$. Therefore
$
d_{1,\min}=\sqrt{\|a_1\|^2-\alpha_1^2}=1.
$
Thus no prescribed deviation $d_1<1$ is feasible: there is no $\lambda\in\mathbb R$ such that $\operatorname{dist}(r_0-\lambda u_1,Y_0)=d_1$. The obstruction comes from the available innovation direction being orthogonal to the residual; dimension lifting restores feasibility by introducing a direction with nonzero alignment with it.
\end{example}


Adding one new orthogonal dimension restores feasibility by allowing a new innovation direction with controlled alignment to $a$. This justifies the need for dimension lifting in order to guarantee universal deviation feasibility and, consequently, collapse of the parametrization error in CCDS.

\begin{theorem}[One-step feasibility by dimension lifting]
\label{thm:feasibility by dimension lifting}
Let $H$ be a real Hilbert space, $Y\subset H$ a closed subspace, and $r\in H$. Define $a := (I-\Pi_Y)r \in Y^\perp$.
Fix a target deviation $d$ satisfying $0 \le d \le \|a\|$.
Then there exists a Hilbert space $\widetilde H := H \oplus \mathbb{R}$,
an isometric embedding $\iota:H\to \widetilde H$, and a unit vector
$u \in (\iota(Y))^\perp \subset \widetilde H$
such that the CCDS distance equation $\operatorname{dist}(\iota(r)-\lambda u,\, \iota(Y)) = d$
admits a real solution $\lambda\in\mathbb{R}$. Moreover, one can choose $u$ so that the minimal achievable deviation is exactly $d_{\min}(u) = d$.
\end{theorem}

\begin{proof}
Define $\widetilde H := H\oplus \mathbb{R}$ with inner product
\[
\langle (x,s),(y,t)\rangle_{\widetilde H} := \langle x,y\rangle_H + st,
\]
and the isometric embedding $\iota(x)=(x,0)$. Then $\iota(Y)$ is a closed subspace and
\[
(\iota(Y))^\perp = Y^\perp \oplus \mathbb{R}.
\]
Let
\[
\widetilde r := \iota(r), \qquad \widetilde a := (I-\Pi_{\iota(Y)})\widetilde r = (a,0),
\]
so that $\|\widetilde a\|=A$.
If $A=0$, then $d=0$ and the claim is trivial. Assume $A>0$ and define
$
e_1 := \frac{\widetilde a}{A}, \ e_2 := (0,1).
$
Then $e_1,e_2\in (\iota(Y))^\perp$ are orthonormal. Choose $\theta\in[0,\pi/2]$ such that
$
\sin\theta = \frac{d}{A}.
$
Define
$
u := \cos\theta\, e_1 + \sin\theta\, e_2 .
$
Then $u\in (\iota(Y))^\perp$ and $\|u\|=1$. Moreover,
$
\alpha := \langle \widetilde a, u\rangle_{\widetilde H}
= A\cos\theta .
$

By the CCDS feasibility formula,
\[
d_{\min}(u) = \sqrt{\|\widetilde a\|^2 - \alpha^2}
= \sqrt{A^2 - A^2\cos^2\theta}
= A\sin\theta
= d .
\]
Hence $d\ge d_{\min}(u)$, and the equation
$
\operatorname{dist}(\widetilde r-\lambda u,\iota(Y)) = d
$
admits a real solution $\lambda$. This completes the proof.
\end{proof}

\paragraph{Resolution of Example \ref{ex:finite_dim_infeasibility} by lifting one dimension.} Embed $H$ isometrically into $\widetilde H = \mathbb{R}^3$
by $x \mapsto (x,0)$, and extend the orthogonal complement with the new basis vector $e_3$. Choose the lifted innovation direction
$\tilde u_t
= \cos\theta \, e_1 + \sin\theta \, e_3,
\ \sin\theta = d_t$. Then $\tilde u_t \in (\iota(Y_{t-1}))^\perp$, $\|\tilde u_t\|=1$, and $\langle \tilde a , \tilde u_t \rangle
= \cos\theta$, $d_{\min}(\tilde u_t)
= \sqrt{1-\cos^2\theta}
= \sin\theta
= d_t$. Hence the CCDS equation
\[
\operatorname{dist}\big(\iota(r_{t-1}(x_1)) - \lambda \tilde u_t ,\, \iota(Y_{t-1})\big) = d_t
\]
admits a real solution, and the deviation $d_t$ becomes feasible after adding a single orthogonal dimension.

\subsection{Stability theorems}
\label{subsec:stability theorems}
\paragraph{Proof of Theorem \ref{thm:stability control interpolation} (Stability of the scaffold/CCDS bottlenecks under relaxed interpolation)}
 \label{app:local-chart-diagnostic}

For each $1\le k\le K$, let $D_k=\{(x_{k, 1}, y_{k,1}), \dots, (x_{k, |D_k|}, y_{k, |D_k|})\}\subseteq \Xi_{\rm {sub}}$ and choose depths $L_{k, 1},\dots, L_{k, |D_k|}$ such that $y_{k, j}=J_k W^{(L_{k,j})}x_{k, j}$ for all $(x_{k,j}, y_{k, j})\in D_k$.
Let \(\mathcal I=\{(k, m, j): 1\le k\le K, \ 0\le m\le d-1,\ 1\le j\le |D_k| \}\) be the fixed finite atom index set of the corresponding scaffold as defined in (\ref{eq:def scaffold}). Write the reference atoms as
$g_i=J_kW^{(m)}x_{k,j}$, with $i=(k,m,j)\in\mathcal I$, and the perturbed atoms as
\(g_i'=J_k'W^{(m)}x_{k,j}\). Let \(G=[g_i]_{i\in\mathcal I}\), \(G'=[g_i']_{i\in\mathcal I}\), \(U=\operatorname{range}(G)\), \(U'=\operatorname{range}(G')\), and \(\eta:=\|G'-G\|_2\). Also set \(V:=\sum_{k=1}^K\operatorname{range}(Y_k)\).

Since \(J_k'=Y_kR_k'\), every perturbed atom \(g_i'\) belongs to \(\operatorname{range}(Y_k)\), hence \(U'\subseteq V\). By the scaffold identity (\ref{eq:U equal range Y}), \(V=U\). Therefore \(U'\subseteq U\).

Let \(r=\operatorname{rank}(G)=\dim U\) and \(\gamma=\sigma_r(G)>0\). Choose \(\delta_0<\gamma/2\). If \(\eta<\delta_0\), then Weyl's singular-value perturbation inequality \cite{Horn:2012} gives \(\sigma_r(G')\ge \sigma_r(G)-\eta>\gamma/2>0\). Hence \(\operatorname{rank}(G')\ge r\). Since \(U'\subseteq U\) and \(\dim U=r\), also \(\operatorname{rank}(G')\le r\). Therefore \(\operatorname{rank}(G')=r\), and so \(U'=U\). It follows immediately that \(\Pi_{U'}=\Pi_U\), and hence
\(\varepsilon_{\rm proj}(f_\Xi,U';\Xi)=\varepsilon_{\rm proj}(f_\Xi,U;\Xi)\).

It remains to prove the parametrization bound. Since \(U'=U\), the projected targets are identical: \(b_\ell:=\Pi_U f_\Xi x_\ell=\Pi_{U'} f_\Xi x_\ell\). Let \(n:=|\Xi_{\mathrm{train}}|\), fix \(\ell\in\{1,\ldots,n\}\), and suppress the index \(\ell\). Let \(r_0=b_\ell\), and let \(r_t\), \(t=0,\dots,T\), be the residuals of the reference CCDS path of the CCDS realizer $\rm CCDS_{\mathcal D}$. At step \(t\), the reference path uses a prescribed atom label \(i_t\), prescribed deviation \(d_t\), and prescribed root branch \(\varepsilon_t\in\{-1,1\}\). Let \(Y_{t-1}:=\operatorname{span}\{u_1,\dots,u_{t-1}\}\), \(P_{t-1}:=\Pi_{Y_{t-1}}\), \(q_t:=(I-P_{t-1})g_{i_t}\), \(u_t:=q_t/\|q_t\|\), \(a_t:=(I-P_{t-1})r_{t-1}\), \(\alpha_t:=\langle a_t,u_t\rangle\), and \(d_{t,\min}^2:=\|a_t\|^2-\alpha_t^2\). The CCDS coefficient is
\(\lambda_t=\alpha_t+\varepsilon_t\sqrt{d_t^2-d_{t,\min}^2}\), and the residual update is \(r_t=r_{t-1}-\lambda_tu_t\).

We transport exactly this finite prescription to the perturbed atoms: the same labels \(i_t\), the same deviations \(d_t\), and the same root branches \(\varepsilon_t\). Thus \(g_{i_t}\) is replaced by the corresponding \(g_{i_t}'\), and all perturbed quantities \(Y_{t-1}'\), \(P_{t-1}'\), \(q_t'\), \(u_t'\), \(a_t'\), \(\alpha_t'\), \(\lambda_t'\), and \(r_t'\) are defined by the same formulas.

We prove by induction that \(\|r_t'-r_t\|\le B_t\eta\) and \(\|u_t'-u_t\|\le A_t\eta\) for constants depending only on the reference trajectory. The base residual estimate is \(\|r_0'-r_0\|=0\), because \(b_\ell'=b_\ell\). Assume the estimates are known up to step \(t-1\). Since the directions \(u_1,\dots,u_{t-1}\) and \(u_1',\dots,u_{t-1}'\) are orthonormal along the CCDS innovation construction, if \(U_{t-1}:=[u_1\ \cdots\ u_{t-1}]\) and \(U_{t-1}':=[u_1'\ \cdots\ u_{t-1}']\), then \(P_{t-1}=U_{t-1}U_{t-1}^\top\) and \(P_{t-1}'=U_{t-1}'(U_{t-1}')^\top\). 
Add and subtract $U_{t-1}(U_{t-1}')^\top$:
$
P_{t-1}'-P_{t-1}
=
(U_{t-1}'-U_{t-1})(U_{t-1}')^\top
+
U_{t-1}\bigl((U_{t-1}')^\top-U_{t-1}^\top\bigr).
$
Taking operator norms,
$
\|P_{t-1}'-P_{t-1}\|_{\rm op}
\le
\|U_{t-1}'-U_{t-1}\|_2\|(U_{t-1}')^\top\|_2
+
\|U_{t-1}\|_2\|(U_{t-1}')^\top-U_{t-1}^\top\|_2.
$
Since $U_{t-1}$ and $U_{t-1}'$ have orthonormal columns, $\|U_{t-1}\|_2=\|U_{t-1}'\|_2=1$.
Thus
$
\|P_{t-1}'-P_{t-1}\|_{\rm op}
\le
2\|U_{t-1}'-U_{t-1}\|_2\le 2\|U_{t-1}'-U_{t-1}\|_F=2 \left(\sum_{s=1}^{t-1}\|u_s'-u_s\|^2\right)^{1/2}$.
Now the induction hypothesis says that for all previous steps $s<t$,
$\|u_s'-u_s\|\le A_s\eta$
for some constants $A_s$ depending only on the reference trajectory. Since there are only finitely many steps, define
$A_{t-1}^{\max}:=\max_{1\le s<t}A_s$.
Then
$\left(\sum_{s=1}^{t-1}\|u_s'-u_s\|^2\right)^{1/2}\le\sqrt{t-1}\,A_{t-1}^{\max}\eta$.
Hence
\begin{equation}
\label{eq:P ineq}
\|P_{t-1}'-P_{t-1}\|_{\rm op}
\le
2\sqrt{t-1}\,A_{t-1}^{\max}\eta
= C_{P,t}\eta.
\end{equation}

Since corresponding atom columns differ by at most \(\eta\), we have \(\|g_{i_t}'-g_{i_t}\|\le \eta\). Therefore
\(q_t'-q_t=(I-P_{t-1}')(g_{i_t}'-g_{i_t})+(P_{t-1}-P_{t-1}')g_{i_t}\), and so
\begin{equation}
\label{eq:q ineq}
\|q_t'-q_t\|\le \eta+C_{P,t}\eta\|g_{i_t}\|\le C_{q,t}\eta.
\end{equation}
By the nondegeneracy assumption, \(\|q_t\|\ge \nu>0\). After decreasing \(\delta_0\), we may assume \(\|q_t'-q_t\|\le \nu/2\). Then by the reverse triangle inequality, $\|q'_t\|\ge\|q_t\|-\|q'_t-q_t\|\ge\nu-\nu/2=\nu/2$. So both $q$ and $q'_t$ stay away from zero.
Now prove the Lipschitz bound. We write $\frac{q'}{\|q'\|}-\frac{q}{\|q_t\|}=\frac{q'_t-q_t}{\|q'_t\|}+q_t\left(\frac{1}{\|q'_t\|}-\frac{1}{\|q_t\|}\right)$. Taking norms, $\left\|\frac{q'_t}{\|q'_t\|}-\frac{q_t}{\|q_t\|}\right\|\le\frac{\|q'_t-q_t\|}{\|q'_t\|}+\|q_t\|\left|\frac{1}{\|q'_t\|}-\frac{1}{\|q_t\|}\right|$.
For the second term, $\left|\frac{1}{\|q'_t\|}-\frac{1}{\|q_t\|}\right|=\frac{\bigl|\|q_t\|-\|q'_t\|\bigr|}{\|q'_t\|\|q_t\|}$. By the reverse triangle inequality, $\bigl|\|q_t\|-\|q'_t\|\bigr|\le\|q_t-q'_t\|$. Therefore $\|q_t\|\left|\frac{1}{\|q'_t\|}-\frac{1}{\|q_t\|}\right|\le\|q_t\|\frac{\|q_t-q'_t\|}{\|q'_t\|\|q_t\|}=\frac{\|q_t-q'_t\|}{\|q_t'\|}$. So $\left\|\frac{q'_t}{\|q'_t\|}-\frac{q_t}{\|q_t\|}\right\|\le\frac{2\|q'_t-q_t\|}{\|q'_t\|}$.
Since $\|q'_t\|\ge\nu/2$, we get $\frac{1}{\|q'_t\|}\le\frac{2}{\nu}$. Hence 
$\|u'_t-u_t\|=\left\|\frac{q'_t}{\|q'_t\|}-\frac{q_t}{\|q_t\|}\right\|\le\frac{4}{\nu}\|q'_t-q_t\|$. And by (\ref{eq:q ineq}) we have
\begin{equation}
\label{eq:u ineq}
\|u'_t-u_t\|\le A_t \eta.
\end{equation}

Now compare the CCDS coefficients. We have
\(a_t'-a_t=(I-P_{t-1}')(r_{t-1}'-r_{t-1})+(P_{t-1}-P_{t-1}')r_{t-1}\). Since the reference trajectory is finite, its residuals are bounded. Together with
(\ref{eq:P ineq}), this gives
\begin{equation}
\label{eq:a eq}
\|a_t'-a_t\|\le C_{a,t}(\|r_{t-1}'-r_{t-1}\|+\eta).
\end{equation}
Also
$\alpha_t'-\alpha_t=\langle a_t'-a_t,u_t'\rangle+\langle a_t,u_t'-u_t\rangle$, hence
\begin{equation}
\label{eq:alpha ineq}
|\alpha_t'-\alpha_t|\le C_{\alpha,t}(\|r_{t-1}'-r_{t-1}\|+\eta).
\end{equation}

Recall $d_{t,\min}^2=\|a_t\|^2-\alpha_t^2$. Similarly, $(d_{t,\min}')^2=\|a_t'\|^2-(\alpha_t')^2$. Then, $(d_{t,\min}')^2-d_{t,\min}^2=\bigl(\|a_t'\|^2-\|a_t\|^2\bigr)-\bigl((\alpha_t')^2-\alpha_t^2\bigr)$.
Taking absolute values, $\left|(d_{t,\min}')^2-d_{t,\min}^2\right|\le\left|\|a_t'\|^2-\|a_t\|^2\right|+\left|(\alpha_t')^2-\alpha_t^2\right|$.
Now $\left|\|a_t'\|^2-\|a_t\|^2\right|=\left|(\|a_t'\|-\|a_t\|)(\|a_t'\|+\|a_t\|)\right|$. Using $|\|a_t'\|-\|a_t\||\le\|a_t'-a_t\|$, we get $\left|\|a_t'\|^2-\|a_t\|^2\right|\le(\|a_t'\|+\|a_t\|)\|a_t'-a_t\|$.
Similarly, $|(\alpha_t')^2-\alpha_t^2|=|\alpha_t'-\alpha_t|\,|\alpha_t'+\alpha_t|$.
Along the finite trajectory, $\|a_t\|$, $\|a_t'\|$, $|\alpha_t|$, and $|\alpha_t'|$ stay bounded for sufficiently small perturbations. Therefore by (\ref{eq:a eq}) and (\ref{eq:alpha ineq}) we have

\begin{equation}
\label{eq:d ineq}
|(d_{t,\min}')^2-d_{t,\min}^2|\le C_{m,t}(\|r_{t-1}'-r_{t-1}\|+\eta).
\end{equation}

Choose the same deterministic root branch in the reference and perturbed problem, for example the minus root: $\lambda_t=\alpha_t-\sqrt{d_t^2-d_{t,\min}^2}$, $\lambda_t'=\alpha'-\sqrt{(d_t)^2-(d_{t,\min}')^2}$.
Then $|\lambda_t'-\lambda_t|\le|\alpha_t'-\alpha_t|+\Big|\sqrt{d_t^2-(d_{t,\min}')^2}-\sqrt{d_t^2-d_{t,\min}^2}\Big|$.
By the feasibility-margin assumption, \(d_t^2-d_{t,\min}^2\ge \eta_0^2>0\). So if the perturbation is small enough that $C_{m, t}(\|r'_{t-1}-r_{t-1}\|+\eta)\le\eta_0^2/2$, then by (\ref{eq:d ineq}) we have $d^2_t-(d_{t,\min}')^2\ge d_t^2-d_{t,\min}^2-|(d_t^2-(d_{t,\min}')^2)-(d_t^2-d_{t,\min}^2)|\ge\eta_0^2-\eta_0^2/2=\eta_0^2/2$. Thus the perturbed distance equation remains strictly feasible.
The map \(s\mapsto \sqrt{s}\) is Lipschitz on \([\eta_0^2/2,\infty)\). Therefore the square-root Lipschitz estimate gives $|\sqrt{(d_t)^2-(d_{t,\min}')^2}-\sqrt{d_t^2-d_{t,\min}^2}|\le C_{\sqrt{\phantom{x}}}|(d_t)^2-(d_{t,\min}')^2-(d_t^2-d_{t,\min}^2)|=C_{\sqrt{\phantom{x}}}\left|(d_{t,\min}')^2-d_{t,\min}^2\right|$.
By (\ref{eq:alpha ineq}) and (\ref{eq:d ineq}) we obtain
\begin{equation}
\label{eq:lambda ineq}
|\lambda_t'-\lambda_t|\le C_{\lambda,t}(\|r_{t-1}'-r_{t-1}\|+\eta).
\end{equation}

Finally, \(r_t'-r_t=(r_{t-1}'-r_{t-1})-(\lambda_t'u_t'-\lambda_tu_t)\), and
\(\lambda_t'u_t'-\lambda_tu_t=(\lambda_t'-\lambda_t)u_t'+\lambda_t(u_t'-u_t)\). Since \(\|u_t'\|=1\), \(\lambda_t\) is bounded along the finite reference trajectory, and by (\ref{eq:u ineq}) and (\ref{eq:lambda ineq}), we get
\(\|r_t'-r_t\|\le L_t\|r_{t-1}'-r_{t-1}\|+D_t\eta\). The induction hypothesis then gives 
\begin{equation}
\label{eq:bounf for residuals}
\|r_t'-r_t\|\le B_t\eta
\end{equation}
for a suitable constant \(B_t\). This closes the induction.

Taking the maximum of the finitely many constants over all samples and all steps, we obtain \(C_{\rm CCDS}>0\) such that \(\|r_{\ell,T_\ell}'-r_{\ell,T_\ell}\|\le C_{\rm CCDS}\eta\) for every \(\ell\). The transported perturbed path is an admissible CCDS realizer $\rm CCDS_{\mathcal D'}$ on \(U'=U\). Therefore
\(\varepsilon_{\rm param}(f_\Xi,U', \rm CCDS_{\mathcal D'})= \frac1n\sum_{\ell=1}^n\|r_{\ell,T_\ell}'\|
\le \frac1n\sum_{\ell=1}^n\|r_{\ell,T_\ell}\|+\frac1n\sum_{\ell=1}^n\|r_{\ell,T_\ell}'-r_{\ell,T_\ell}\|
\le \varepsilon_{\rm param}(f_\Xi,U, \rm CCDS_{\mathcal D})+C_{\rm CCDS}\eta\).
Recalling that \(\eta=\|G'-G\|_2\) gives the desired bound.

How to choose $\delta_0$ explicitly: For each trajectory \(\ell\) and step \(t\), the positive nondegeneracy and feasibility margins in the preceding induction provide radii
\(\delta_{1,\ell,t}>0\) and \(\delta_{2,\ell,t}>0\) for which the normalization and root-feasibility arguments remain valid. Indeed,
for each trajectory \(\ell\) and step \(t\), let
$
\mu_{\ell,t}
:=
d_{\ell,t}^{\,2}-d_{\ell,t,\min}^{\,2}>0
$
be the reference feasibility margin. Display (\ref{eq:q ineq}) gives
$
\|q_{\ell,t}'-q_{\ell,t}\|
\leq C_{q,\ell,t}\eta.
$
Define
\[
\delta_{1,\ell,t}
:=
\frac{\nu}{2C_{q,\ell,t}}.
\]
Then \(\eta<\delta_{1,\ell,t}\) implies
\(\|q_{\ell,t}'-q_{\ell,t}\|<\nu/2\), preserving the nonzero normalization.
After the induction estimate, (\ref{eq:bounf for residuals}) gives
\[
\|r_{\ell,t-1}'-r_{\ell,t-1}\|
\leq B_{\ell,t-1}\eta
\]
and by \eqref{eq:d ineq} we obtain
\[
\left|
(d_{\ell,t,\min}')^2-d_{\ell,t,\min}^2
\right|
\leq
C_{m,\ell,t}(B_{\ell,t-1}+1)\eta.
\]
Therefore define
\[
\delta_{2,\ell,t}
:=
\frac{\mu_{\ell,t}}
{2C_{m,\ell,t}(B_{\ell,t-1}+1)}.
\]
Then \(\eta<\delta_{2,\ell,t}\) implies
\[
d_{\ell,t}^{\,2}-(d_{\ell,t,\min}')^2
\geq
\frac{\mu_{\ell,t}}{2}>0,
\]
so the perturbed square-root equation remains feasible.

 Define
\[
\delta_j
:=
\min_{1\leq\ell\leq n}
\min_{1\leq t\leq T_\ell}
\delta_{j,\ell,t},
\qquad j=1,2,
\]
and set
\[
\delta_0:=\min\{\gamma/2,\delta_1,\delta_2\}.
\]
For a zero-step trajectory, the inner minimum is taken to be \(+\infty\).
\hfill\(\blacksquare\)

\paragraph{Proof of Theorem~\ref{thm:observable_sparse_stability} (observable sparse-stability theorem)}
\label{app:observable_sparse_stability}

We first establish a quantitative comparison between the orbit interpolants $J_k$ and the canonical local-chart operators $A_k$, then transfer this comparison to their dictionaries and the corresponding constrained sparse approximation problems.

\medskip
\noindent
\textit{Step 1: Comparison of the local interpolation operators.}

Fix a block $D_k$. By assumption,
\[
\ker X_k\subseteq\ker Y_k.
\]
Since $X_k^\dagger X_k$ is the orthogonal projector onto $(\ker X_k)^\perp$, this inclusion implies
\[
Y_k=Y_kX_k^\dagger X_k.
\]
Indeed, every $c\in\mathbb R^{s_k}$ admits an orthogonal decomposition
\[
c=c_0+c_1,
\qquad
c_0\in\ker X_k,
\quad
c_1\in(\ker X_k)^\perp.
\]
Consequently,
\[
Y_kX_k^\dagger X_kc
=
Y_kc_1
=
Y_k(c_0+c_1)
=
Y_kc,
\]
because $Y_kc_0=0$.

With $A_k:=Y_kX_k^\dagger$, we therefore have
$
A_kX_k=Y_k.
$
In particular,
\[
A_kx_{k,j}=y_{k,j},
\qquad j=1,\ldots,s_k.
\]

Moreover,
\[
\|A_k\|
=\|Y_kX_k^\dagger\|
\leq \|Y_k\|_F\|X_k^\dagger\|
=\frac{\|Y_k\|_F}{\sigma_{\min}^{+}(X_k)}.
\]

Recall that $Z_k$ has full column rank and
$
E_k=\operatorname{range}(Z_k).
$
Hence
\[
Z_kZ_k^\dagger=\Pi_{E_k}.
\]
Using $J_k=Y_kZ_k^\dagger=A_kX_kZ_k^\dagger$, we obtain
\[
J_k-A_k\Pi_{E_k}
=A_kX_kZ_k^\dagger-A_kZ_kZ_k^\dagger
=A_k(X_k-Z_k)Z_k^\dagger.
\]

Taking operator norms gives
\[
\begin{aligned}
\|J_k-A_k\Pi_{E_k}\|
&\leq
\|A_k\|\|X_k-Z_k\|_F\|Z_k^\dagger\|\\
&\leq
\frac{\|Y_k\|_F}
{\sigma_{\min}^{+}(X_k)\sigma_{\min}(Z_k)}
\|X_k-Z_k\|_F.
\end{aligned}
\]
Since
\[
\|X_k-Z_k\|_F
=
\left(
\sum_{\ell=1}^{s_k}
\|x_{k,\ell}-z_{k,\ell}\|^2
\right)^{1/2},
\]
we conclude that
\begin{equation}
\label{eq:local_chart_perturbation}
\|J_k-A_k\Pi_{E_k}\|
\leq
\frac{\|Y_k\|_F}
{\sigma_{\min}^{+}(X_k)\sigma_{\min}(Z_k)}
\left(
\sum_{\ell=1}^{s_k}
\|x_{k,\ell}-z_{k,\ell}\|^2
\right)^{1/2}.
\end{equation}

\medskip
\noindent
\textit{Step 2: Control by the observable partition score.}

For each block, set
\[
\delta_k:=
\frac{\|Y_k\|_F}
{\sigma_{\min}^{+}(X_k)\sigma_{\min}(Z_k)}
\left(
\sum_{\ell=1}^{s_k}
\|x_{k,\ell}-z_{k,\ell}\|^2
\right)^{1/2}.
\]
Then \eqref{eq:local_chart_perturbation} reads
\[
\|J_k-A_k\Pi_{E_k}\|\leq\delta_k.
\]

By the definition of $S(\mathcal P)$,
\[
\begin{aligned}
NS(\mathcal P)
&=
\sum_{k=1}^{K}\sum_{j=1}^{s_k}
\frac{\|Y_k\|_F}{\sigma_{\min}^{+}(X_k)}
\operatorname{dist}(x_{k,j},E_k)\\
&\quad+
\sum_{k=1}^{K}\sum_{j=1}^{s_k}
\frac{\|Y_k\|_F\|x_{k,j}\|}
{\sigma_{\min}^{+}(X_k)\sigma_{\min}(Z_k)}
\left(
\sum_{\ell=1}^{s_k}
\|x_{k,\ell}-z_{k,\ell}\|^2
\right)^{1/2}\\
&=
\sum_{k=1}^{K}\sum_{j=1}^{s_k}
\frac{\|Y_k\|_F}{\sigma_{\min}^{+}(X_k)}
\operatorname{dist}(x_{k,j},E_k)
+
\sum_{k=1}^{K}\delta_k
\sum_{j=1}^{s_k}\|x_{k,j}\|.
\end{aligned}
\]

All terms in the first sum are nonnegative. Therefore,
\begin{equation}
\label{eq:partition_controls_chart}
\sum_{k=1}^{K}
\delta_k\sum_{j=1}^{s_k}\|x_{k,j}\|
\leq NS(\mathcal P).
\end{equation}

In particular, for every block $D_k$,
\begin{equation}
\label{eq:individual_chart_control}
\delta_k\sum_{j=1}^{s_k}\|x_{k,j}\|
\leq NS(\mathcal P).
\end{equation}

\medskip
\noindent
\textit{Step 3: Stability of the orbit dictionary atoms.}

Fix an arbitrary index
\[
r=(k,j,m)\in\mathcal I.
\]
The corresponding atoms in $\mathcal D$ and $\mathcal D_{\mathrm{chart}}$ are
\[
g_r=J_kW^m x_{k,j},
\qquad
\widetilde g_r=A_k\Pi_{E_k}W^m x_{k,j}.
\]
Their difference is
\[
g_r-\widetilde g_r
=
(J_k-A_k\Pi_{E_k})W^m x_{k,j}.
\]
Hence
\[
\|g_r-\widetilde g_r\|
\leq
\|J_k-A_k\Pi_{E_k}\|
\|W^m x_{k,j}\|
\leq
\delta_k\|W^m x_{k,j}\|.
\]

By the definition of $\kappa(\mathcal P)$,
\[
\|W^m x_{k,j}\|
\leq
\kappa(\mathcal P)
\sum_{\ell=1}^{s_k}\|x_{k,\ell}\|.
\]
Therefore,
\[
\|g_r-\widetilde g_r\|
\leq
\kappa(\mathcal P)
\delta_k
\sum_{\ell=1}^{s_k}\|x_{k,\ell}\|
\leq
N\kappa(\mathcal P)S(\mathcal P),
\]
where the final inequality follows from
\eqref{eq:individual_chart_control}.

Since $r\in\mathcal I$ was arbitrary,
\begin{equation}
\label{eq:uniform_atom_stability}
\max_{r\in\mathcal I}
\|g_r-\widetilde g_r\|
\leq
N\kappa(\mathcal P)S(\mathcal P).
\end{equation}

This estimate controls the perturbation of every atom in the complete orbit dictionary, not only those corresponding to the selected interpolation depths.

\medskip
\noindent
\textit{Step 4: Stability of bounded-coefficient sparse synthesis.}

Let $a\in\mathbb R^P$ satisfy
\[
|\{r\in\mathcal I:a_r\neq0\}|\leq s,
\qquad
\sum_{r\in\mathcal I}|a_r|\leq B.
\]
The corresponding realizations are
\[
S_{\mathcal D}(a)=\sum_{r\in\mathcal I}a_rg_r,
\qquad
S_{\mathcal D_{\mathrm{chart}}}(a)
=\sum_{r\in\mathcal I}a_r\widetilde g_r.
\]
Consequently,
\[
\begin{aligned}
\|S_{\mathcal D}(a)
-S_{\mathcal D_{\mathrm{chart}}}(a)\|
&=
\left\|
\sum_{r\in\mathcal I}
a_r(g_r-\widetilde g_r)
\right\|\\
&\leq
\sum_{r\in\mathcal I}
|a_r|\|g_r-\widetilde g_r\|\\
&\leq
\left(\sum_{r\in\mathcal I}|a_r|\right)
\max_{r\in\mathcal I}
\|g_r-\widetilde g_r\|\\
&\leq
BN\kappa(\mathcal P)S(\mathcal P),
\end{aligned}
\]
by \eqref{eq:uniform_atom_stability}.

Thus
\begin{equation}
\label{eq:sparse_synthesis_stability}
\|S_{\mathcal D}(a)
-S_{\mathcal D_{\mathrm{chart}}}(a)\|
\leq
BN\kappa(\mathcal P)S(\mathcal P)
\end{equation}
uniformly over every coefficient vector satisfying the prescribed constraints.

\medskip
\noindent
\textit{Step 5: Stability of optimal sparse realization errors.}

Fix $y\in\mathbb R^C$, and write
\[
\mathcal A_{s,B}
:=
\left\{
a\in\mathbb R^P:
|\{r\in\mathcal I:a_r\neq0\}|\leq s,\ 
\sum_{r\in\mathcal I}|a_r|\leq B
\right\}.
\]

For every $a\in\mathcal A_{s,B}$, the reverse triangle inequality and \eqref{eq:sparse_synthesis_stability} give
\[
\begin{aligned}
\big|
\|y-S_{\mathcal D}(a)\|
-
\|y-S_{\mathcal D_{\mathrm{chart}}}(a)\|
\big|
&\leq
\|S_{\mathcal D}(a)
-S_{\mathcal D_{\mathrm{chart}}}(a)\|\\
&\leq
BN\kappa(\mathcal P)S(\mathcal P).
\end{aligned}
\]

In particular,
\[
\|y-S_{\mathcal D}(a)\|
\leq
\|y-S_{\mathcal D_{\mathrm{chart}}}(a)\|
+
BN\kappa(\mathcal P)S(\mathcal P)
\]
for every $a\in\mathcal A_{s,B}$.

Taking the infimum over this common admissible set yields
\[
e_{s,B}(y;\mathcal D)
\leq
e_{s,B}(y;\mathcal D_{\mathrm{chart}})
+
BN\kappa(\mathcal P)S(\mathcal P).
\]

Interchanging the roles of the two dictionaries gives
\[
e_{s,B}(y;\mathcal D_{\mathrm{chart}})
\leq
e_{s,B}(y;\mathcal D)
+
BN\kappa(\mathcal P)S(\mathcal P).
\]

Combining the two inequalities, we conclude that
\[
\left|
e_{s,B}(y;\mathcal D)
-
e_{s,B}(y;\mathcal D_{\mathrm{chart}})
\right|
\leq
BN\kappa(\mathcal P)S(\mathcal P).
\]

Since $y$, $s$, and $B$ were arbitrary, the assertion follows.
\hfill\(\blacksquare\)

\section{Experimental configurations}
\label{sec:experiment-config}

\subsection{Data splits, preprocessing, and repetition}

Gas Sensor Drift batches are concatenated in batch order and split at 80\%: training subsets are sampled from the first part, and the final 20\% is held out. ISOLET uses the first 6,238 examples for training and the remaining 1,559 for testing. MNIST uses the standard 60,000-example training pool, with the first 48,000 forming the training pool and the last 12,000 held out, kept untouched during model fitting and shared by all methods. MNIST pixels are divided by 255 before preprocessing.

Standardization, randomized PCA (random state 0), and subsequent coordinate standardization are fitted on the selected training subset and then applied to held-out data. Gas at 128 features omits PCA. The main runs use subset seeds 13--17 and construction seeds 1000--1004. The Gas and ISOLET heads use the corresponding construction seed; the MNIST head uses seed 1000 in all five runs. Consequently, the reported standard deviations concern the specified repeated training and construction procedure, not independent test splits.

This matched preprocessing protocol is essential for interpretation: differences among the feature-based methods cannot be attributed to different input representations or access to additional training information. The experiments therefore compare how each method uses the same compressed geometry, while the orbit method additionally exposes the role of scaffold discovery and constructive realization.

\begin{table}[t]
\centering
\small
\begin{tabular}{lrrrrr}
\hline
Setting & Training size & PCA variance (\%) & Anchor budget (\%) & Anchors & Atoms \\
\hline
Gas, 25 & 4,000 & 99.2 & 1.0 & 40 & 1,000 \\
Gas, 128 & 8,000 & 100 & 0.7 & 56 & 7,168 \\
ISOLET, 25 & 6,238 & 70.6 & 5.0 & 311 & 7,775 \\
ISOLET, 100 & 6,238 & 88.8 & 5.0 & 311 & 31,100 \\
MNIST, 25 & 5,000 & 46.0 & 3.0 & 150 & 3,750 \\
MNIST, 100 & 10,000 & 74.8 & 0.7 & 70 & 7,000 \\
\hline
\end{tabular}
\caption{Main construction settings. PCA variance is approximate and varies slightly with the selected subset; 100\% for raw Gas denotes omission of PCA. Anchor counts are the actual integer counts. An anchor budget limits dictionary construction: all selected training labels are used for target projection, realization, and prediction.}
\label{tab:main-data-config}
\end{table}

The anchor budget is a discovery budget, not the amount of data available to the final predictor. After the scaffold is selected, all designated training targets are projected onto that scaffold, realized by Phase~2, and used to train the Phase~3 coefficient head. This distinction separates structural discovery from subsequent coefficient learning.

\subsection{Construction and coefficient prediction}

The main predictive runs fix \(w_j=0.5^{1/d}\), so the cycle product is 0.5; these runs do not learn the shift weights. Up to five disjoint farthest-first anchor candidates are considered. A greedy orbit-independence test uses threshold 0.1, followed by splitting large blocks. Target block sizes are \(\lceil d/3\rceil\) for PCA Gas and MNIST, \(\lceil d/12\rceil\) for raw Gas, and \(\lceil d/7\rceil\) for ISOLET. Candidate selection uses training projection error, with an early stopping threshold of \(10^{-7}\). The local chart score uses normalized scalar class indices as a partition heuristic; the block interpolants used for synthesis use one-hot class targets. 

Interpolants use a pseudoinverse of the regularized normal matrix with ridge \(10^{-6}\) in float32. The synthesis dictionary contains all \(d\) recorded depths of each anchor under its own block interpolant, with \(P=d|\Xi_{\mathrm{sub}}|\). Phase 2 primarily uses the trivial-reference distance-equation solver with \(\gamma=0\), alignment threshold 0.99, batch size 256, and an iteration cap of \(P\). Its stopping threshold is \(10^{-7}\|\Pi_U y\|+10^{-10}\). Final certificate support in the main summaries counts coefficients with absolute value greater than \(10^{-8}\).

The primary Phase~2 configuration deliberately uses the minimal-deviation boundary, making it a residual-collapse reference and explaining the near-zero realization errors reported above. The interior CCDS experiments are separate diagnostics: they retain the same dictionary while testing the additional path-control freedom created by prescribing nonminimal feasible deviations.

The global head has two hidden layers of the same width, GELU activations and LayerNorm, with dropout 0.15 after the first hidden layer. Its linear output has \(P\) coefficient coordinates. For a fit set \(F\), the implemented objective is
\[
\begin{aligned}
\mathcal{L}_F(\theta)
=
\frac{1}{|F|}
\sum_{i\in F}
\Bigl[
&\operatorname{CE}(G\alpha_\theta(x_i),y_i)
+\lambda_{\mathrm{coeff}}\|\alpha_\theta(x_i)-\beta_i\|_2 \\
&+\frac{\lambda_{\mathrm{out}}}{C}
\|G\alpha_\theta(x_i)-c_i\|_2^2
+\frac{\lambda_1}{P}\|\alpha_\theta(x_i)\|_1
\Bigr].
\end{aligned}
\]

The coefficient discrepancy is an unsquared Euclidean norm. AdamW uses batches of 128. After construction and realization on the full selected training set, 15\% is reserved for head validation and 85\% for head fitting. Validation therefore selects the extension head conditional on an already fitted representation. Checkpoints are evaluated every 25 epochs and at the final epoch, using the validation objective without its \(\ell_1\) term. Validation and test coefficients with magnitude at most \(10^{-5}\) are set to zero before synthesis.

The coefficient head is therefore supervised by the constructive certificate itself, while the output-consistency term keeps its synthesized prediction aligned with the realized target. The resulting predictor is not an unconstrained classifier attached to the scaffold; it is an extension of the sparse orbit-coordinate rule produced by Phase~2.

\begin{table}[t]
\centering
\small
\begin{tabular}{lrrr}
\hline
Setting & Hidden width & Learning rate & Epochs \\
\hline
Gas, 25 & 53 & \(10^{-3}\) & 125 \\
Gas, 128 & 126 & \(10^{-5}\) & 500 \\
ISOLET, 25 & 196 & \(6\times10^{-4}\) & 75 \\
ISOLET, 100 & 286 & \(4\times10^{-5}\) & 250 \\
MNIST, 25 & 96 & \(10^{-3}\) & 250 \\
MNIST, 100 & 256 & \(10^{-4}\) & 400 \\
\hline
\end{tabular}
\caption{Main coefficient-head configurations. Weight decay and \((\lambda_{\mathrm{coeff}},\lambda_{\mathrm{out}},\lambda_1)\) are \(10^{-2}\) and \((0,1,10^{-2})\) for Gas, \(10^{-3}\) and \((0,2,0)\) for ISOLET, and \(10^{-2}\) and \((10^{-2},1,0)\) for MNIST.}
\label{tab:main-head-config}
\end{table}

\subsection{Synthetic geometry and realization diagnostics}
In the weak alignment and matched rank tests we use 900 synthetic inputs of dimension 10, seven classes, generator seed 4, and a 500-example training subset. Inputs have noisy class directions and sample-wise rotations. A normalized random readout \(B\) maps the operator response to seven outputs; the operator \(C\) has logarithmically spaced singular values from 1 to \(10^{-4}\) and spectral norm one. Readout seed is 456 and operator seed is 103. The block dictionary is constructed from one-hot labels obtained from these responses, whereas projection and realization diagnostics evaluate the continuous target \(BCx\).
The weak-alignment run uses 35 anchors, 14 blocks, and 350 atoms. Shift weights are optimized for 100 Adam steps at learning rate \(10^{-3}\), starting from \(0.5^{1/d}\). The nested solver uses seven innovations and alignment threshold 0.8. On weakly aligned steps, \(d_t=\max\{d_{t,\min},\gamma\|a_t\|\}\), clipped at \(\|a_t\|\); other steps use the boundary up to numerical tolerance. The minus root is used. The matched-rank test uses ten anchor seeds 101--110 and twenty random references at each rank one through six. Selection uses the leading singular direction of the training responses compressed to the remaining candidate space.

The operator-family test uses a near-parallel union of six three-dimensional subspaces in 35 dimensions, with class perturbation scale 0.02 and noise scale 0.05. Generator seed is 4. A 2,000-example training subset is standardized and reduced to 15 PCA coordinates, retaining approximately 98.2\% of variance. A single random 60-anchor construction has 30 two-anchor blocks and 900 atoms. Its shift is fitted for 100 Adam steps at \(10^{-3}\), initialized at one; the resulting dictionary is reused across all eight response operators. The families are an ill-conditioned SVD operator, identity, orthogonal, rank-five, spiked, sparse, Toeplitz, and an anisotropic diagonal operator.

The nested solver uses six innovations, alignment threshold 0.8, and \(\gamma=0.05\). If \(b=10^{-3}\|\Pi_U y\|\) and \(B_{t-1}\) is the accumulated squared coordinate discrepancy, its weak-alignment step allocates
\[
\delta_t=\min\left\{\gamma|\alpha_t|,
\sqrt{\frac{\max\{0,b^2-B_{t-1}\}}{6-t+1}}\right\},
\qquad d_t^2=d_{t,\min}^2+\delta_t^2.
\]

\subsection{Dictionary structure and interpolation variants}
The waveform task uses four Fourier motifs in dimension 32 with seven frequencies, random cyclic shifts, amplitudes in \([0.7,1.3]\), and shared random frequency-wise rotations of the targets. Each seed 101--110 generates 720 fit examples and 900 held-out examples, split into 150 validation and 750 test examples. Each 20- or 28-anchor dictionary uses one block, fixed cycle product 0.5, and ridge \(10^{-6}\). The unstructured arm draws Gaussian directions within the orbit dictionary's numerical span, using relative singular-value threshold \(10^{-5}\), matches each row norm, and preserves the selected-depth anchor rows.

Three greedy updates form the training realization targets. Both proxy heads have two width-128 GELU/LayerNorm layers and dropout 0.15, and are trained for 200 epochs with AdamW, learning rate \(10^{-3}\), weight decay \(10^{-2}\), and batch size 128. The proxy loss is target MSE plus 0.25 times MSE to the three-step realization. Checkpoints minimize held-out validation MSE. Predictions use two MP updates on the proxy output. The direct MLP uses the same architecture and optimization settings with target MSE alone.

The interpolation comparison has 15,000 synthetic examples with five informative Gaussian coordinates, 44 correlated nuisance coordinates, and one mildly aligned spurious feature. The split is 12,000 training and 3,000 clean-test examples. Clean labels are the sign of the informative linear score. Exactly 0\%, 20\%, or 40\% of training labels are flipped, with corruption seed \(42+1000q\) at fraction \(q\); test labels are unchanged. Both arms use PCA dimension 10, 120 anchors, five paired subset seeds 13--17, and a width-51 global head trained for 300 epochs at \(10^{-3}\), with weight decay \(10^{-2}\) and \((\lambda_{\mathrm{coeff}},\lambda_{\mathrm{out}},\lambda_1)=(1,1,0.1)\). The exact arm uses an unregularized pseudoinverse with a constructive correction fallback. The relaxed arm uses ridge \(10^{-3}\). Each arm optimizes shift weights for 100 Adam steps at \(10^{-3}\).

\subsection{Discovery-budget configurations and values}

Both sweeps use subset seed 13 and construction seed 1000, with one run per budget. Gas uses 3,000 training examples and ten PCA coordinates, retaining approximately 95.0\% of variance. MNIST uses 5,000 training examples and twenty PCA coordinates, retaining approximately 41.9\%. All heads use batches of 128 and train for 300 epochs, with \(\lambda_{\mathrm{out}}=1\).

For Gas, the hidden width is 51 and weight decay is \(10^{-2}\). Budgets 1\%, 3\%, and 5\% use learning rate \(10^{-3}\) and \(\lambda_{\mathrm{coeff}}=1\); 7\% uses the same rate and \(\lambda_{\mathrm{coeff}}=0.2\); 10\% and 15\% use \(10^{-4}\) and \(\lambda_{\mathrm{coeff}}=0\). All use \(\lambda_1=0.1\). Shift optimization uses 100 Adam steps at \(10^{-3}\), initialized at \(0.5^{1/d}\).

For MNIST, learning rate is \(10^{-3}\) throughout. Budgets 1\%, 3\%, and 5\% use width 64, weight decay \(10^{-2}\), \(\lambda_{\mathrm{coeff}}=1\), and \(\lambda_1=0.1\). At 7\%, \(\lambda_{\mathrm{coeff}}=0.15\). At 10\%, width increases to 156, weight decay decreases to \(10^{-3}\), and both coefficient tracking and \(\ell_1\) penalties are removed. Shift learning uses 25 Adam steps at \(10^{-3}\), relative-weight penalty 0.01, and
\[
w_j=\exp\left(\frac{\log(0.5)}{d}+\theta_j-\frac1d\sum_{k=0}^{d-1}\theta_k\right),
\]
which fixes the cycle product at 0.5.

\subsection{Baselines}

The common baselines use logistic regression with at most 1,000 iterations, a 100-tree random forest, an RBF SVM with \(C=1\) and scale-based bandwidth, and histogram gradient boosting with learning rate 0.1. The MLP uses the corresponding head's two hidden widths, ReLU, learning rate \(10^{-3}\), a 10\% early-stopping validation fraction, and at most 2,000 iterations. It is not a complete architecture or parameter-count match. Remaining unspecified parameters use the notebook's library defaults.

ISOLET additionally uses a Nystr\"om RBF feature map with 800 components, bandwidth parameter 0.002, and ridge classifier regularization 0.01, and a distance-weighted 15-nearest-neighbor classifier. Their accuracies are \(91.20\pm0.05\%\) and \(86.79\pm0.00\%\) at 25 features, and \(92.89\pm0.20\%\) and \(87.94\pm0.00\%\) at 100 features. Gas additionally reports CORAL logistic regression at \(63.64\pm2.97\%\) and \(69.31\pm1.19\%\). The recorded delay-ridge reference reaches \(65.36\pm3.78\%\) and \(73.86\pm2.65\%\), exceeding WBSNN in the raw-feature setting; however, it forms training delays after random subset ordering, so it is not a chronological temporal comparison. The corresponding Hankel/LDA reference reaches \(34.13\pm5.89\%\) and \(49.93\pm1.90\%\).

The MNIST CNN uses two \(3\times3\) convolutional layers with 16 and 32 channels, batch normalization, ReLU, and \(2\times2\) max pooling, followed by a width-128 hidden layer and dropout 0.5. It is trained on raw images for seven epochs with Adam, learning rate \(10^{-3}\), and batch size 32.



\end{document}